%% file: main.tex
\documentclass[conference]{IEEEtran}
\IEEEoverridecommandlockouts
\usepackage{cite}
\usepackage{amsmath,amssymb,amsfonts}
\usepackage{algorithmic}
\usepackage{graphicx}
\usepackage{textcomp}
\usepackage{multirow}
\usepackage{xcolor}
\usepackage{subfigure}
\usepackage{pifont}
\usepackage[ruled,linesnumbered]{algorithm2e}
\def\BibTeX{{\rm B\kern-.05em{\sc i\kern-.025em b}\kern-.08em
    T\kern-.1667em\lower.7ex\hbox{E}\kern-.125emX}}
    
\begin{document}

\title{GitFL: Adaptive Asynchronous Federated Learning  using Version Control}

\author{\IEEEauthorblockN{Ming Hu\IEEEauthorrefmark{2},
Zeke Xia\IEEEauthorrefmark{3},
Zhihao Yue\IEEEauthorrefmark{3},
Jun Xia\IEEEauthorrefmark{3},
Yihao Huang\IEEEauthorrefmark{2},
Yang Liu\IEEEauthorrefmark{2}, and
Mingsong Chen$^*$\IEEEauthorrefmark{3}}
\IEEEauthorblockA{\IEEEauthorrefmark{2}Nanyang Technological University, Singapore}
\IEEEauthorblockA{\IEEEauthorrefmark{3}East China Normal University, Shanghai, China}
\thanks{$^*$Corresponding Author.}
}








\maketitle

\input{abstract}

\input{intro}

\input{related}

\input{approach}

\input{experiment}

\input{conclusion}

\end{document}

%% file: abstract.tex
\begin{abstract}

As a promising distributed machine learning paradigm that enables  
collaborative training without compromising  data privacy, 
Federated Learning (FL) has been increasingly used in 
 AIoT (Artificial Intelligence of Things) design.
However,  due to the
lack of efficient management of 
straggling devices, 
existing FL methods greatly
suffer from the problems of 
low inference accuracy and
long training time. 
Things become even worse when taking 
various uncertain factors (e.g., network delays, 
performance variances
caused by process variation) existing
in AIoT scenarios into account.  
To address this issue, this paper proposes a novel asynchronous FL framework named GitFL, whose implementation is
inspired by 
the famous
version control system Git.  
Unlike traditional FL,  the cloud server of GitFL maintains 
a master model (i.e., the global model) together with 
a set of branch models indicating 
the trained 
local models committed by  selected 
devices, where the master model is updated
based on both all the pushed  branch
models and their version information, and only the branch models after the pull operation are dispatched to devices.
By using 
our proposed Reinforcement Learning (RL)-based  
device selection mechanism, a pulled
branch model  with an
older version  will be more likely 
to  be dispatched to a
faster and less frequently selected device for the next round of 
local training. 
In this way, GitFL enables both effective 
control of model staleness and adaptive
load balance of versioned models among 
straggling devices, thus avoiding the performance deterioration. 
%
Comprehensive experimental results on well-known models and datasets show that, compared with state-of-the-art asynchronous FL methods, GitFL can  achieve up to 2.64X training acceleration
and  7.88\% inference accuracy improvements in various uncertain scenarios.
\end{abstract}

\begin{IEEEkeywords}
AIoT, Federated Learning, Asynchronous Aggregation, Uncertainty, Reinforcement Learning, Version Control
\end{IEEEkeywords}

%% file: intro.tex
\section{Introduction}

Along with the prosperity
of Artificial Intelligence of Things (AIoT), 
we are Federated Learning (FL) ~\cite{chand_dac2022,zhang_tcad2021} is becoming more and more popular in the design of 
AIoT applications, e.g., autonomous driving, industrial automation, and control \cite{lin_aspdac2022}, commercial surveillance, and healthcare systems \cite{wu_iccad2021}. 
As an promising
collaborative  learning paradigm, 
FL adopts a cloud-based architecture, where AIoT devices  focus on local  training  based on their own data  while the cloud server is responsible for both knowledge aggregation and dispatching. 
Although FL  enables knowledge sharing among AIoT devices without compromising data privacy, 
existing FL methods 
inevitably
suffer  from the problems of low inference performance and
long overall training time. 
Generally,  a typical AIoT system involves a variety of heterogeneous devices, which have different computation and communication capabilities (e.g., CPU/GPU frequencies, memory size, network bandwidths) and specified  workloads (e.g., the number  of data samples, data distributions).
In this case, devices with weaker  performance  (a.k.a., {\it stragglers})
will take much longer local training time than the other selected devices. 
This is quite harmful to 
the most widely-used synchronous  FL methods such as Federated Average (FedAvg) \cite{fedavg},
which requires that  the cloud server should
wait for the responses from all the selected devices before aggregation. 
According to the ``wooden barrel theory'', when applying 
synchronous  FL methods , the training  time of 
one FL communication
round will be determined by the slowest selected devices. 
As a result, the existence of stragglers 
will significantly slow down the process of overall FL training.

To mitigate the shortcomings caused by stragglers in 
synchronous FL, various asynchronous FL methods
(e.g., FedAsync~\cite{xie2019asynchronous}, HADFL \cite{cao_dac2021} and Helios~\cite{xu_dac2021})  have been investigated. 
Unlike synchronous FL, in  each   training
round the asynchronous cloud server performs an update of the global model once it receives a local model from some device.
However, due to different participating
frequencies of devices in asynchronous updating, the stale local 
models trained  by 
 stragglers 
can easily result in poor
training quality, especially for non-IID (Independent Identically Distribution) scenarios. 
As a trade-off between  synchronous and asynchronous FL methods, 
by  forcing to  
synchronize multiple stale local models in one training round, semi-asynchronous FL methods (e.g., SAFA~\cite{safa} and FedSA~\cite{fedsa})  can be used to partially alleviate the model 
staleness
issue. However, such
semi-asynchronous FL  methods also bring   new problems. 
For example, due to the 
large staleness discrepancies, the cloud server may frequently synchronize with 
some stragglers before they complete their local training. 
In this case, the stragglers may not have a chance to participate
in global model updating, thus biasing
the training of global models. 
Thing becomes even worse when taking various uncertain factors in AIoT scenarios into account, since both  asynchronous and semi-asynchronous FL methods assume  that
the performance metrics (e.g., network delay and computation capacities) of 
AIoT devices are fixed. However, this is not true in practice. For example, 
due to  process variation during the manufacturing of microelectronic circuits, 
the performance of devices with the same type varies significantly \cite{pv}. 
Due to the over-pessimistic estimation of the performance metrics,  
the potential of asynchronous FL is strongly suppressed. 
Clearly, due to the lack of efficient management mechanisms for
stale models within an uncertain environment, 
it is hard for existing asynchronous FL methods to quickly converge to a higher inference performance.

\begin{figure*}[ht] 
	\begin{center} 
		\includegraphics[width=0.9\textwidth]{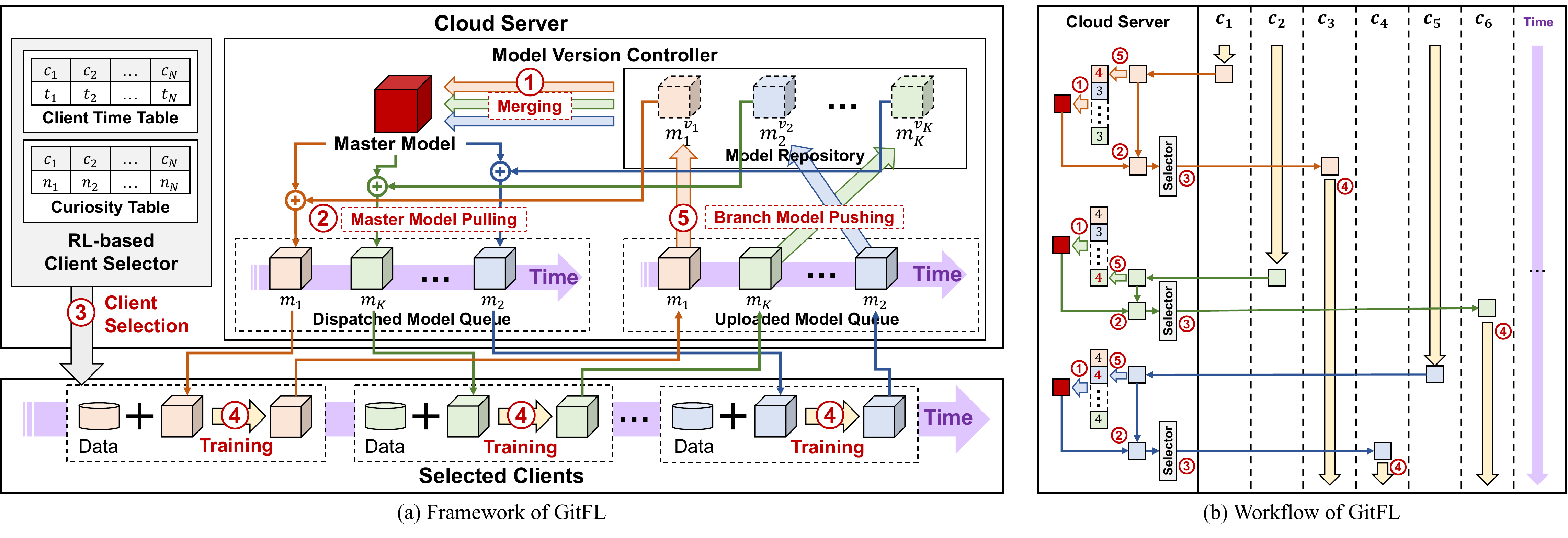}
  \vspace{-0.05 in}
		\caption{Framework and workflow of GitFL.}
  \vspace{-0.2 in}
		\label{fig:framework} 
	\end{center}
\end{figure*}

Inspired by the implementation of Git, a famous version control system, 
this paper proposes
a novel asynchronous FL approach named GitFL, which enables  
 efficient
management of straggler devices to accommodate various uncertainties in an adaptive manner. 
Unlike traditional FL methods, GitFL  does not dispatch its global model directly  to 
its selected devices for local training. Instead, it keeps a set of branch models that can traverse
 different devices asynchronously, and dispatches new branch
 models pulled from the global model to devices. 
 Here, the global model is only used to accumulate the versioned knowledge obtained 
 by the branch models. 
To support the adaptive load balance of version models on straggler devices,  GitFL is  apt to  dispatch 
a pulled  branch model with an older version to a faster 
and less frequently  device for the 
next round of local training.
In summary, this paper makes the following three major contributions:
\begin{itemize}
\item
Based on the push/pull operations, we established a 
novel Git-like asynchronous FL framework, which 
supports both effective management of straggler devices
and version control of their stale models within an uncertain environment. 


\item 
We develop a new Reinforcement Learning (RL)-based 
curiosity-driven exploration method to  support the wise selection of straggler devices in an adaptive manner. Our approach enables the  load balance  of  versioned models among AIoT devices, thus avoiding  the performance deterioration caused by stragglers.


\item 
We conduct experiments on both well-known datasets and models to show the effectiveness of our approach in terms of 
convergence rate and inference performance for both IID and non-IID scenarios.

\end{itemize}

The rest of this paper is organized as follows. Section~\ref{sec:relatedwork} introduces  the backgrounds and related work  of existing  FL methods. Section~\ref{sec:approach}  details the implementation of  our GitFL approach. Section~\ref{sec:experiment} gives the experimental results. Finally, Section~\ref{sec:conclusion} concludes this paper.

%% file: related.tex
\section{Backgrounds and Related Work}\label{sec:relatedwork}
\subsection{Preliminaries to FL}

The traditional FL framework consists of a central cloud server and multiple clients.
The cloud server maintains a global model for FL training.
In each FL training round, the cloud server dispatches the global model to multiple clients.
Then, each client uses its local data for local training and uploads the trained model to the cloud server.
Finally, the cloud server generates a new global model by aggregating all the uploaded models.
Typically, the model aggregation is based on FedAvg \cite{fedavg}, which is defined as follows:
\begin{equation}
\begin{split}
&\min_{w} F(w) = \frac{1}{K}\sum_{k = 1}^{K} f_k(w), \\
\ s.t., &\ f_k(w) = \frac{1}{|d_k|} \sum_{i = 1}^{|d_k|} \ell (w, \langle x_i,l_i \rangle),
\end{split}
\end{equation}
where $K$ is the number of clients that participate in local training, $d_k$ is the dataset of the $k^{th}$ client, the $|d_k|$ denotes the number of data samples in $d_k$, $\ell$ denotes the customer-defined loss function (e.g., cross-entropy loss), $\langle x_i,y_i \rangle$ denotes a sample $x_i$ with its label $l_i$.

\subsection{FL on Heterogeneous Devices}
To deal with the straggler issue caused by heterogeneous devices, there are numerous optimization FL methods based on synchronous and asynchronous architectures.
For example, in~\cite{xu_dac2021},  Xu {\it et al.} proposed a synchronous FL method Helios, which accelerates the training speed of stragglers by adaptive reducing model parameters.
In~\cite{xie2019asynchronous}, Xie {\it et al.} proposes an asynchronous federated learning framework FedAsync, in which the cloud server aggregates the uploaded model with the global model in real time according to the staleness of nodes.
In~\cite{safa}, Wu {\it et al.} proposes a semi-asynchronous FL method SAFA, which uses a cache structure to bypass uploaded models according to their versions for model aggregation.
Similarly, in~\cite{fedsa}, Ma {\it et al.} sets a model buffer for model aggregation to achieve semi-asynchronous FL.
Although these methods can alleviate the problem of stragglers, they do not consider uncertain environments.


To the best of our knowledge, GitFL is the first attempt to use a version control mechanism for asynchronous FL and an RL-based client selection strategy to enable the load balance of versioned models among AIoT devices and adaption to uncertain environments.
Compared with state-of-the-art FL methods, our GitFL approach can achieve the best test convergence rate and inference performance for both IID and
non-IID scenarios.

%% file: approach.tex
\section{Our GitFL Approach}\label{sec:approach}
\subsection{Overview}

Figure~\ref{fig:framework} presents the framework and workflow of our GitFL approach, which consists of a central cloud server and multiple clients. 
The cloud server contains two key components, i.e., the model version controller and the RL-based client selector.
The model version controller updates the master model and branch models asynchronously according to the version of each branch model.
The RL-based client selector chooses clients for local training according to the number of past selections, historical training time, and the version of the dispatched branch model.

The model version controller includes a model repository to store all the branch models with its version, an uploaded queue to receive trained branch models, and a dispatched queue for assigning updated branch models. 
%
The RL-based client selector consists of a client time table and a count table, respectively. The client time table records the training time of each client and the count table records the selected number of each client. 
In detail, to perform timekeeping for each model, it starts when the target model is assigned to local clients from the dispatched model queue. After the model is trained in local clients and pushed to the uploaded model. The update of tables is in real-time.
With the help of these two tables, the client selector can prompt all clients adequately participate in training, thus reducing the version difference between branch models.
Please note that the selection strategy is according to the version of a target branch model and the two tables which record the operating efficiency of local clients. The branch model with a higher version is preferred to assign to a client with a longer training time, and vice versa.
For a model with a medium version, the selector prefers to select the client with less number of selections. 

Each iteration of a branch model in GitFL includes five key steps: i) model merging, the controller updates the dispatched branch model by merging it with the new master model and pushing it into the dispatched model queue;
ii) model pulling, the controller updates the dispatched branch model by merging it with the new master model and pushing it into the dispatched model queue; iii) client selection, the client selector selects a client for branch model dispatching; iv) local training, the selected client trains the received branch model using its local data; and v) model pushing, the controller pushes the branch model into the repository and updates its version.

\setlength{\textfloatsep}{5pt}
\begin{algorithm}[th]
\caption{Implementation of  GitFL}
\label{alg:gitfl}
\scriptsize
\KwIn{
    {\bf i)} $T$,  training time;
    {\bf ii)} $C$, the set of clients;
    {\bf iii)} $K$, \# of clients  participating in local training.
 }
 \KwOut{
   $m_g$, the global model.
 }
\textbf{GitFL($T$,$C$,$K$)} \Begin{
$R\leftarrow [m_1, m_2,...,m_{K}]$\;\label{line:init_start}
$V[k]\leftarrow 0$ for $k \in [1,K]$\;\label{line:init_v}
$T_{c}[i],T_{t}[i]\leftarrow 0$ for $i \in [1,|C|]$\;\label{line:init_tab}
$T_{start}\leftarrow Time()$\;\label{line:init_end}
/*parallel for*/\\
\For{$i$ = 1, ..., $K$}{ \label{line:train_start}
    \While{$Time()-T_{start}<T$}{\label{line:train_model_start}
        $M\leftarrow ${\it Merging}$(R,V)$; // Model Merging\\\label{line:merge}
        $m_i\leftarrow ${\it ModelPull}$(i, V, M, R[i])$; // Model Pulling\\\label{line:model_pull}
        $c\leftarrow ${\it ClientSel}$(i, V, T_c, T_t, C)$; // Client Selection\\\label{line:client_selection}
        $t_{start}\leftarrow Time()$\;\label{line:train_time_start}
        $m_i\leftarrow ${\it LocalTrain}$(c,m_i)$; // Model Training\\\label{line:local_training}
        $t_{end}\leftarrow Time()$\;\label{line:train_time_end}
        $t\leftarrow t_{end} - t_{start}$\;\label{line:train_time}
        $V[i] \leftarrow V[i] + 1$\;\label{line:version_update}
        $R[i]\leftarrow m_i$; // Model Pushing\\\label{line:model_push}
        $T_c[c]\leftarrow T_c[c] + 1$\;\label{line:count_table_update}
        $T_t[c]\leftarrow \frac{[(T_c[c]-1) \times T_t[c] + t]}{T_c[c]}$\;\label{line:time_table_update}
    }\label{line:train_model_end}
}\label{line:train_end}
$m_g\leftarrow $Merging$(R,V)$\;\label{line:global_model_generation}
{\bf return} $m_g$\;
}
\end{algorithm}
\vspace{-0.1in}

\subsection{GitFL Implementation}
Algorithm~\ref{alg:gitfl} presents the implementation of our GitFL approach.
We assume that there are at most $K$ clients participating in local training at the same time, which means that GitFL adopts $K$ branch models for local training.
Line \ref{line:init_start} initializes the model repository $R$, which is a list to store all the branch models.
Line \ref{line:init_v} initializes a list $V$, which records the version of all the branch models.
Line \ref{line:init_tab} initializes the count table $T_c$ and the client time table $T_t$.
Line \ref{line:init_end} sets the function {\it Time()} to record the start time of FL training and is used to get the timestamp in real time.
Lines \ref{line:train_start}-\ref{line:train_end} presents the FL training process of $K$ branch models, where the ``for'' loop is a parallel loop.
Lines \ref{line:train_model_start}-\ref{line:train_model_end} present the detail of each branch model training process.
In Line \ref{line:merge}, the function {\it Merging()} merges all the branch models in $R$ to generate a master model $M$.
In Line \ref{line:model_pull}, the function {\it ModelPull()} pulls down the master model to aggregate with the $i^{th}$ branch model according to the version list $V$, where $m_i$ is the aggregated model.
In Line \ref{line:client_selection}, the function {\it ClientSel()} selects a client according to $T_c$, $T_t$, and the version of a dispatched branch model.
Line \ref{line:local_training} dispatches the branch model to the selected client for local training.
In Line \ref{line:train_time_start}, $t_{start}$ uses the timestamp to record the dispatching operation of a branch model.
In Line \ref{line:train_time_end}, $t_{end}$ uses the timestamp to record the receiving operation of a branch model. 
In Line \ref{line:train_time}, $t$ records  the time of a whole local training process of a specified branch model, which includes the time of network communication and local model training.
After receiving the branch model, Line \ref{line:version_update} updates its version and Line \ref{line:model_push} pushes it to the model repository.
In addition, Line \ref{line:count_table_update} and Line \ref{line:time_table_update} update $T_c$ and $T_t$, respectively.

\subsection{Model Version Control}
GitFL adopts a model version control strategy to guide knowledge sharing between the master model and branch models, which consists of three key operations, i.e., model merging, model pulling, and model pushing.
To implement version control, GitFL sets a model repository that stores all branch models and their version information.

\subsubsection{Model Merging}
GitFL generates the master model by merging all branch models in the model repository. Since the branch model with a lower version may reduce the performance of the master model, GitFL gives each branch model a weight based on its version to guide the merging. The merging process is shown as follows:
\begin{equation}\label{eq:master}
Merging(R,V)= \frac{\sum^K_{k=1}{R[k]\times V[k]}}{\sum^K_{k=1}{V[k]}}
\end{equation}
where $R$ denotes the model repository, $K$ presents the number of branch models, and $V$ is the list of version information. 
Here, we directly use  model versions as the merging weights instead of the difference of versions.
At the beginning of training, models with a higher version should dominate the merging, since they are more accurate.
At this time, either model versions  or differences of versions are both reasonable to be used as weights.
However, along with the progress of FL training, all the models are becoming well-trained. At this time, we expect all the models are adequately participating in the merging. Compared with differences of versions, model versions are more likely to equally merge the models.
%

\subsubsection{Model Pulling}
To enable knowledge sharing  across branch models, each branch model needs to pull down the master model before model dispatching. 
In each pulling step, GitFL updates a branch model by aggregating it with the master model, and the aggregation weight depends on the version of this branch model. Here, a branch model with a higher version will be assigned a larger weight. 
This is because, on the one hand, the models with higher versions are more likely to be well-trained, which means they only need less knowledge from the master model. On the other hand, the knowledge of stragglers contained in the master model may have negative effects on the models with higher versions.
For a model with a lower version, due to insufficient training, it needs to learn more knowledge from the master model. Therefore, it is assigned a fewer weight in aggregation. 
Based on the above consideration,
GitFL uses a version control variable to guide the aggregation as follows:
\begin{equation}
v^i_{ctrl} = V[i] - \frac{1}{K}\sum^K_{k=1}V[k]
\end{equation}
where $i$ denotes the index of a branch model. $v^i_{ctrl}$ denotes the difference between the version of $i^{th}$ branch model and the average version of all branch models. 
By using the version control variable, the model-pulling process is defined as follows:
\begin{equation}
{\it ModelPull}(i, V, M, m_i)= \frac{max(10+v^i_{ctrl},2)\times m_i + M}{max(10+v^i_{ctrl},2) + 1}
\end{equation}
where $M$ is the master model generated by Equation \eqref{eq:master}.
Note that the excessive weight of the master model in model pulling makes the update of a branch model too coarse-grained, which may lead to the gradient divergence problem. 
Therefore, we set the weight of each branch model greater than that of the master model, which enables the branch model to acquire knowledge from the master model more smoothly and fine-grained.

\subsubsection{Model Pushing}
When the uploaded queue receives a trained branch model, the model version controller pushes this received branch model to the model repository and updates its version information.
As shown in Figure~\ref{fig:framework}(b), after receiving the model $m_K$, the model repository updates the version of $m_K$ from $3$ to $4$.
In addition,
the model repository will replace the old version of a model with the new one.
Note that for each branch model, the model repository only reserves the latest version. Therefore, in GitFL, the model repository only stores $K$ models.

\subsection{RL-based Client Selection}
Although GitFL uses the master model as a bridge to implement asynchronous training of branch models, due to the uncertainty of clients and network, the version gaps between branch models lead to a low-performance problem. 
To balance the versions of branch models, we design a reinforcement learning strategy to select clients based on the branch model version, the historical training time of clients, and the selection number of clients.

\subsubsection{Problem Definition}
In GitFL, the client selection process can be defined as a Markov Decision Process (MDP), which is presented as a four-tuple $M=\langle \mathcal{S},\mathcal{A},\mathcal{F},\mathcal{R}\rangle$ as follows:
\begin{itemize}
\item $\mathcal{S}$ is a set of states. We use a vector $s=\langle B_{m}, V, C_t, T_c\rangle$ to denote the state of GitFL, where $B_{m}$ denotes the set of branch models that waits for dispatching, $V$ is the list of version information for all the branch models, $C_t$ indicates the set of clients for current local training, and $T_c$ is a table that records the selection number of all the clients.
\item $\mathcal{A}$ is a set of actions. At the state of $s=\langle i, V\rangle$, the action $a$ aims to dispatch the $i^{th}$ branch model to the selected client. The selection space of $\mathcal{A}$ is all the clients in $C$.
\item $\mathcal{F}$ is a set of transitions. It records the transition $s\overset{a}{\longrightarrow}s'$ with the action $a$.
\item $\mathcal{R}$ is the reward function. Here, we combine the selection number and training time as the reward to evaluate the quality of a selection.
\end{itemize}

\subsubsection{Version- and Curiosity-Driven Client Selection}
GitFL adopts a version- and curiosity-driven strategy to evaluate rewards. The client selector prefers to select a client with a higher reward.

The version-driven strategy is used to balance the version differences between branch models. 
For the branch model with a lower version, we prefer to dispatch it to an efficient client (i.e., less training time) and vice versa.
GitFL uses a client time table $T_t$ to record the average historical training time of each client.
When the $i^{th}$ branch model is waiting for dispatching, the version reward for the client $c$ is measured as follows:
\begin{equation}\label{eq:version}
R_v(c,i) = (V[i]-\frac{\sum^K_{k=1} V[k]}{K})(\frac{T_t[c] - \frac{\sum^{|C|}_p T_t[p]}{|P|}}{max(T_t)}),
\end{equation}
where $|C|$ denotes the number of clients, $T_t[c]$ indicates the average historical training time of $c$. The difference between the model version and the average version determines the scope of $R_v$. Note that, the client selection is independent of client training time when the version of a branch model is equal to the average version.

To ensure the adequacy of client selection, we exploit curiosity-driven exploration~\cite{pathak2017curiosity} as one of the strategies to evaluate rewards, while the client with a less number of selections will get a higher curiosity reward.
GitFL uses the count table $T_c$ to record the selection number of each client and 
uses Model-based Interval Estimation with Exploration Bonuses (MBIE-EB)~\cite{bellemare2016nips} to measure the curiosity reward for the client $c$ as follows:
\begin{equation}\label{eq:curiosity}
R_c(c,i) = \frac{1}{\sqrt{T_c[c]}},
\end{equation}
where $T_c[c]$ indicates the selection number of $c$. Note that the initialize value of $T_c[c]$ is 1. The reward of each client is measured by combining $R_v$ and $R_c$ as follows:
\begin{equation}\label{eq:reward}
R(c,i) = max(0,R_v(c,i)+R_c(c,i)).
\end{equation}
If the difference between the model version and the average version is high, the reward will be mainly determined by the version reward. 
If the model version is close to the average version, the reward will be mainly determined by the curiosity reward. 
Note that the minimum value of a reward is 0. If the reward of a client equals 0, the client will not be selected.
Otherwise, based on the above reward definition, the client selector chooses a client for model dispatching with the probability defined as follows:
\begin{equation}\label{eq:pro}
P(c,i) = \frac{R(c,i)}{\sum^{|C|}_{j=1} R(j,i)}.
\end{equation}

%% file: experiment.tex
\section{Performance Evaluation}\label{sec:experiment}

To evaluate  the effectiveness of our approach, we implemented GitFL using the PyTorch framework.
We compared our GitFL method with both classical FedAvg \cite{fedavg} and three state-of-the-art 
FL methods (i.e., FedAsync \cite{xie2019asynchronous}, SAFA \cite{safa}, and FedSA \cite{fedsa}), where the former two are asynchronous and the latter two are 
semi-asynchronous.
For SAFA and FedSA, we set the model buffer size to half the number of active clients.
To enable fair comparison, we used an SGD optimizer with a
learning rate of 0.01 and a momentum of 0.5 for both  all the baselines and  GitFL. For each client,
we set the batch size to 50 and performed five epochs for each local training.
All the experimental results were obtained from an Ubuntu workstation with an Intel i9 CPU, 64GB memory, and an NVIDIA RTX 3090 GPU.

\subsection{Experimental Settings}

\subsubsection{Settings of Uncertainties}
We considered two kinds of uncertainties, i.e., 
the uncertainty of devices 
computation time  caused by process variation \cite{pv}, 
and the uncertainty caused by network delay. 
In the experiment, we assumed that both of them 
follow the Gaussian distributions. 
Table~\ref{tab:config} shows the settings of both uncertainties, 
which are simulated in the FL training processes of experiments. 
Here, we assume that there exist five kinds of devices and five communication 
channels. Generally, devices and communication channels 
with higher  quality have better computation and communication performance with lower variances as specified in the table.


\begin{table}[h]
\centering
\vspace{-0.1in}
\caption{Settings for Devices and Their Communication}
\scriptsize
\begin{tabular}{|c|c|c|}
\hline
\multirow{1}{*}{Quality} & Device Settings & Network Settings\\
\hline
\hline
Excellent & $N(100,5)$ & $N(10,1)$ \\
\cline{1-3}
High & $N(150,10)$ & $N(15,2)$ \\
\cline{1-3}
Medium & $N(200,20)$ & $N(20,3)$ \\
\cline{1-3}
Low & $N(300,30)$ & $N(30,5)$ \\
\cline{1-3}
Critical & $N(500,50)$ & $N(80,10)$ \\
\hline
\end{tabular}
\label{tab:config}
\end{table}

\begin{table*}[t]
\centering
\caption{Comparison of Test accuracy  for both IID and non-IID scenarios}
\vspace{-0.05in}
\scriptsize
\begin{tabular}{|c|c|c|c|c|c|c|}
\hline
\multirow{2}{*}{Dataset} & Heterogeneity & \multicolumn{5}{c|}{Test Accuracy (\%)} \\
\cline{3-7}
 & Settings & FedAvg~\cite{fedavg} & FedAsync~\cite{xie2019asynchronous} &  SAFA~\cite{safa} & FedSA~\cite{fedsa} & GitFL (Ours)\\
\hline
\hline
\multirow{4}{*}{CIFAR-10} & $\alpha=0.1$& $42.87\pm 5.32$& $46.48\pm 2.14$& $41.21\pm 3.42$ & $46.97\pm 3.33$ & ${\bf 54.85\pm 0.92}$\\
                    & $\alpha=0.5$ & $61.07\pm 0.48$& $62.40\pm 0.51$& $56.46\pm 0.89$ & $62.56\pm 0.32$ &  ${\bf 70.36\pm 0.20}$\\
                    & $\alpha=1.0$ & $63.42\pm 0.63$& $65.08\pm 0.29$& $59.75\pm 0.29$ & $66.55 \pm 0.16$ &  ${\bf 72.65\pm 0.18}$\\
                    & IID & $64.80\pm 0.13$& $64.89\pm 0.12$& $62.72\pm 0.19$ & $65.82\pm 0.15$ &  ${\bf 72.40\pm 0.15}$\\
\hline
\multirow{4}{*}{CIFAR-100} & $\alpha=0.1$ & $34.41\pm 0.45$& $34.88\pm 0.67$& $30.42\pm 0.57$ & $35.77\pm 0.46$ &  ${\bf 40.61\pm 0.28}$\\
                       & $\alpha=0.5$ & $41.76\pm 0.32$& $42.82\pm 0.32$& $37.61\pm 0.76$ & $42.53\pm 0.40$ &  ${\bf 47.63\pm 0.22}$\\
                       & $\alpha=1.0$ & $42.08\pm 0.23$& $43.08\pm 0.67$& $38.90\pm 0.33$& $43.32\pm 0.17$ &  ${\bf 48.86\pm 0.23}$\\
                       & IID & $43.04\pm 0.16$& $43.40\pm 0.31$& $40.84\pm 0.22$ & $43.39\pm 0.14$ &  ${\bf 49.26\pm 0.14}$\\
\hline
\multirow{1}{*}{FEMNIST} & $-$ & $82.58\pm 0.32$& $82.29\pm 0.91$& $74.68\pm 1.40$ & $82.63\pm 0.36$ &  ${\bf 83.17\pm 0.10}$\\
\hline
\end{tabular}
\label{tab:acc}
\vspace{-0.1in}
\end{table*}

\begin{table*}[t]
\centering
\caption{Comparison of Training Time and Communication Overhead  to Achieve Given Target Inference Accuracy}
\scriptsize
 \vspace{-0.1in}
\begin{tabular}{|c|c|c|c|c|c|c|c|c|c|c|c|}
\hline
Heterogeneity & \multirow{2}{*}{ Acc. (\%)} & \multicolumn{2}{c|}{FedAvg \cite{fedavg}} & \multicolumn{2}{c|}{FedAsync \cite{xie2019asynchronous}} & \multicolumn{2}{c|}{SAFA \cite{safa}} & \multicolumn{2}{c|}{FedSA \cite{fedsa}} & \multicolumn{2}{c|}{GitFL} \\
\cline{3-12}
 Settings & & Time & Comm. &  Time & Comm. &  Time & Comm. &  Time & Comm. &  Time & Comm. \\
\hline
\hline
\multirow{2}{*}{ $\alpha = 0.1$} &
 40 & 162203 & 3050 &  71566 & 2610 & 231008 & 6809 & {\bf 35067} & {\bf 1018} & 45770 & 1880 \\
 \cline{2-12}
& 45 & - & - &  237357 & 8645 &  - & - &  191477 & 5538 &  {\bf 72418} & {\bf 3000} \\
\hline

\multirow{2}{*}{ $\alpha = 0.5$} &
 55 & 40267 & 740 &  34400  & 1250 & 158434 & 4675 & 20532 & {\bf 608} & {\bf 16884} & 640 \\
  \cline{2-12}
& 60 & 214392 & 3990 &  114673 & 4155 &  - & - &  103915 & 3033 &  {\bf 24192} & {\bf 950} \\
\hline
\multirow{2}{*}{ $\alpha = 1.0$} &
 59 & 46584 & 880 &  25608  & 945 & 172888 & 5103 & 20774 & {\bf 608} & {\bf 16685} & 630 \\
  \cline{2-12}
& 64 & 269044 & 5050 &  140430 & 5135 &  - & - &  85616 & 2493 &  {\bf 25641} & {\bf 1010} \\
\hline
\multirow{2}{*}{ IID} &
60 & 19738 & 360 &  13091 & 445 &  75217 & 2203 &  13375 & {\bf 408} &  {\bf 12081} & 440 \\
 \cline{2-12}
& 64 & 43489 & 800 &  27196 & 1030 &  - & - &  22476 & 678 &  {\bf 15131} & {\bf 570} \\
\hline
\end{tabular}
\label{tab:time}
\vspace{-0.1in}
\end{table*}

\subsubsection{Settings of Datasets and Models}

We compared the performance of GitFL and four baselines on 
three well-known datasets, i.e., CIFAR-10, CIFAR-100 and FEMNIST~\cite{leaf}. 
To investigate the performance of GitFL 
on non-IID scenarios,
we adopted the Dirichlet distribution $Dir(\alpha)$  to control the heterogeneity of client data when using  CIFAR-10 and CIFAR-100 datasets, where 
 a smaller value of $\alpha$ indicates a higher data heterogeneity on devices.
 Note that we did not apply $Dir(\alpha)$ on FEMNIST dataset, since 
 FEMNIST itself 
 is naturally non-IID
distributed.
For the experiments on datasets CIFAR-10 and CIFAR-100, we 
 we assumed each of them  involves 100 AIoT devices in total and in each FL round there were only 10 of them  selected
for local training  at the same time. 
However, for datateset FEMNIST, there are a total of 180 devices and each FL round 
involves 18  devices selected for local training. 
Moreover, to show the pervasiveness  of our GitFL framework, we also take
three DNN models (i.e., CNN, ResNet-18 and VGG-16) with difference 
sizes and structures into account.

\subsection{Performance Comparison}

\subsubsection{Comparison of Accuracy}
Table~\ref{tab:acc} compares  the test accuracy 
between  GitFL and all the baselines on three datasets with different 
non-IID and IID settings using ResNet-18 model. 
%
From this table,
we can find that GitFL can achieve the highest accuracy for all the cases.
As an example of  dataset CIFAR-10, when $\alpha=0.1$, 
GitFL can achieve an  improvement of 7.88\%   over the best inference result obtained by 
FedSA.

\begin{figure}[th]
\vspace{-0.15in}
	\centering
	\subfigure[$\alpha=0.1$]{
		\centering
		\includegraphics[width=0.14\textwidth]{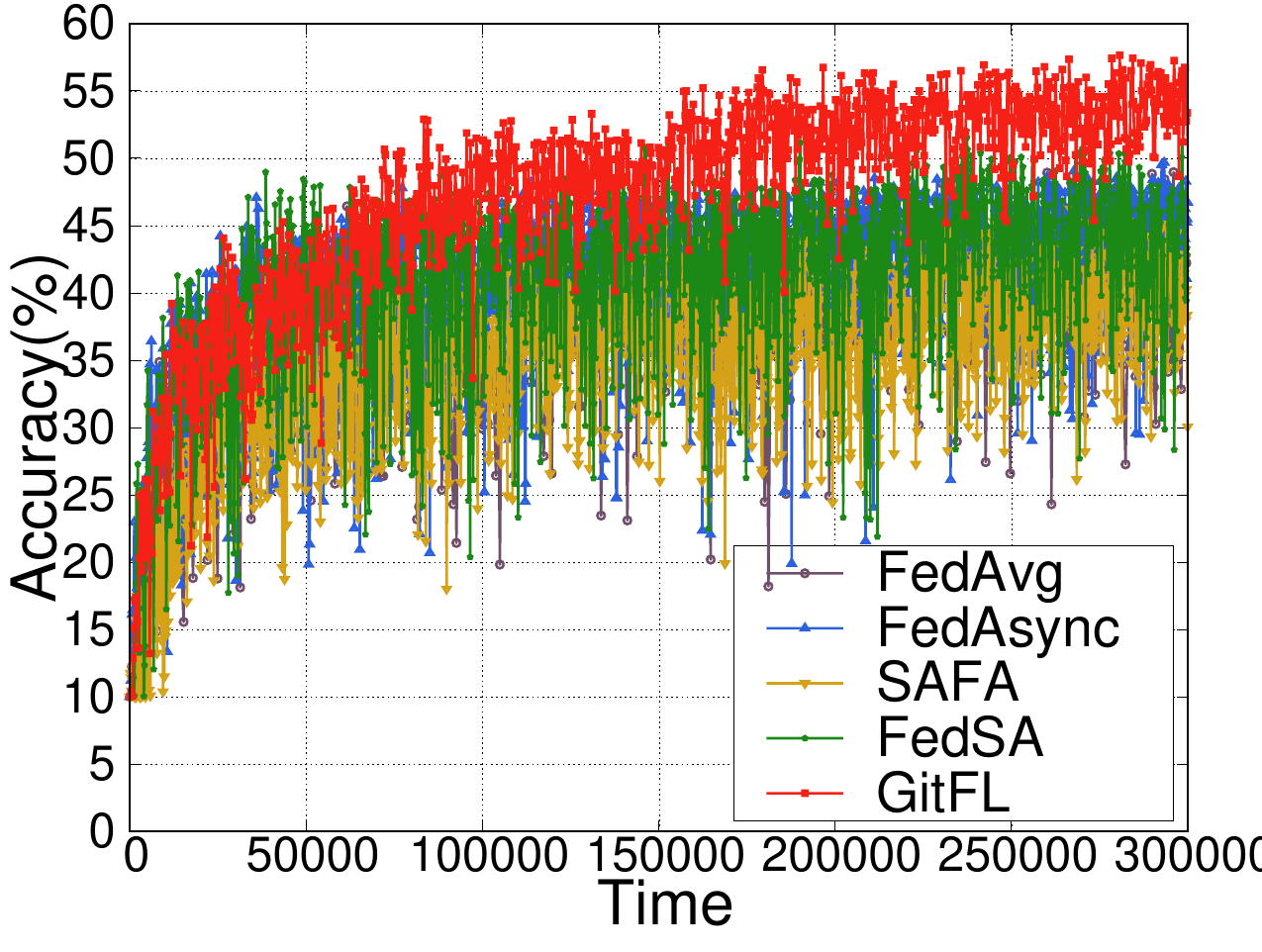}
		\label{fig:cifar-10-0.1}
	}
	\subfigure[$\alpha=0.5$]{
		\centering
		\includegraphics[width=0.14\textwidth]{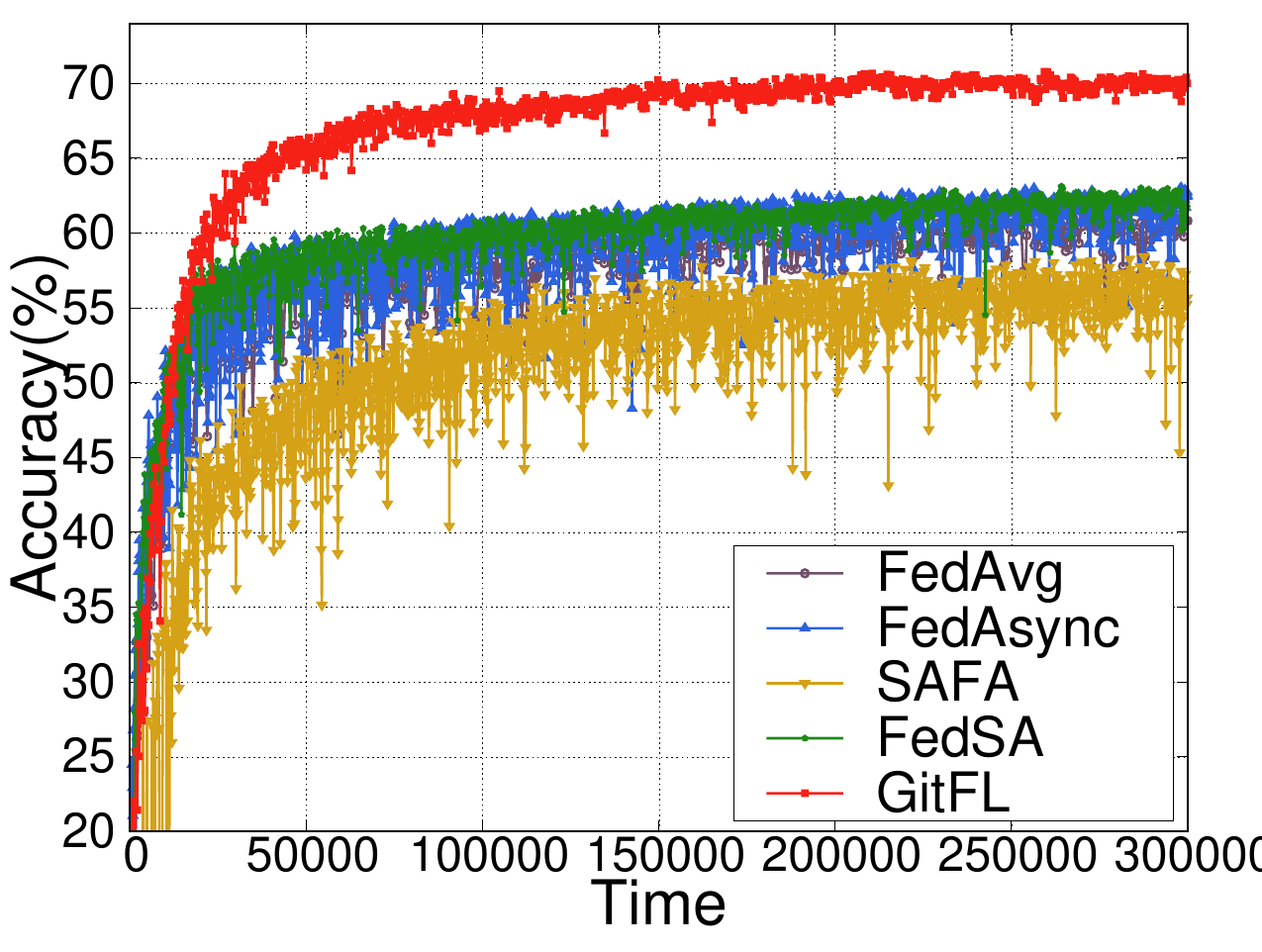}
		\label{fig:cifar-10-0.5}
	}
	\\ \vspace{-0.1in}
	\subfigure[$\alpha=1.0$]{
		\centering
		\includegraphics[width=0.14\textwidth]{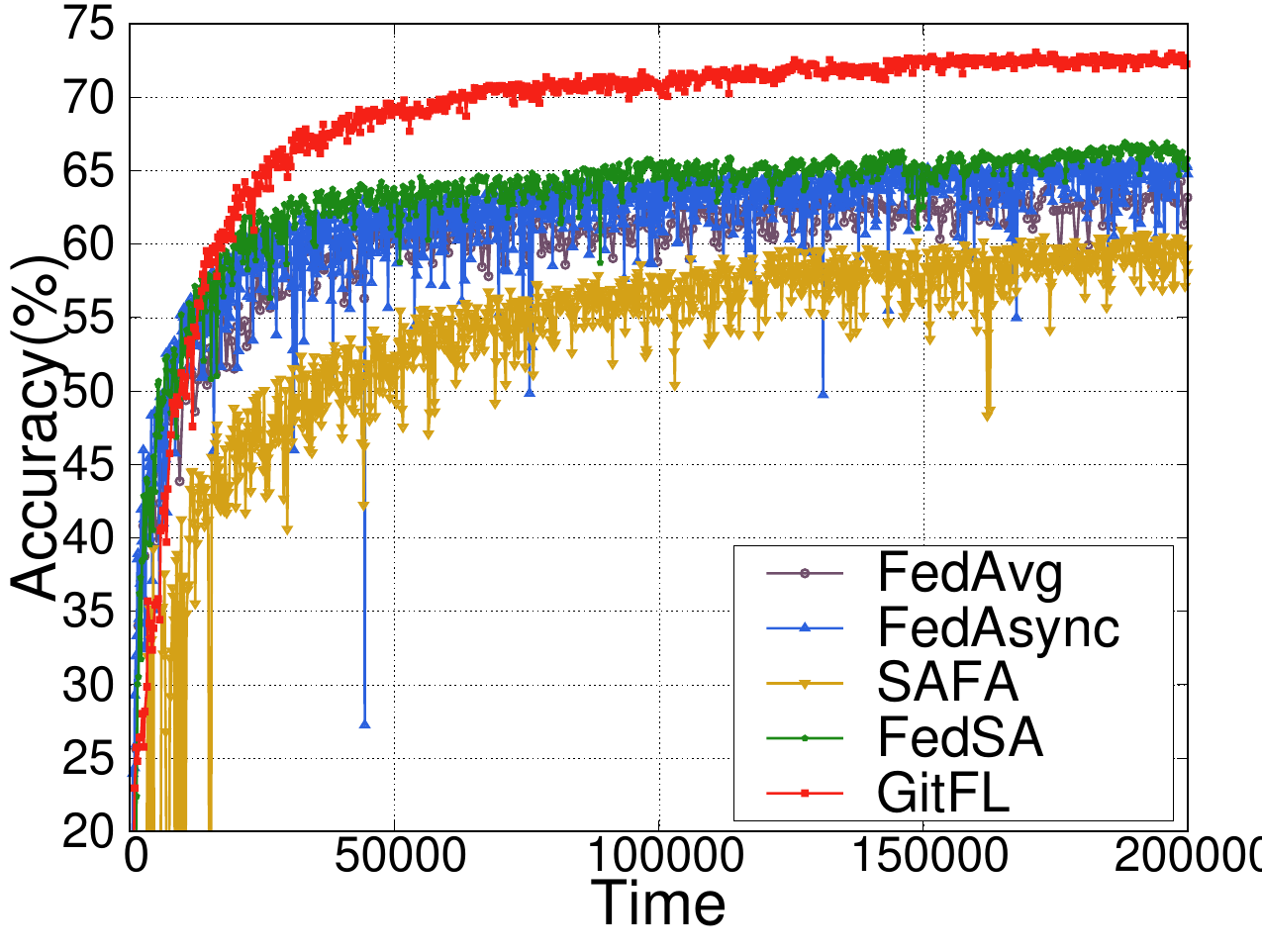}
		\label{fig:cifar-10-1.0}
	}
	\subfigure[IID]{
		\centering
		\includegraphics[width=0.14\textwidth]{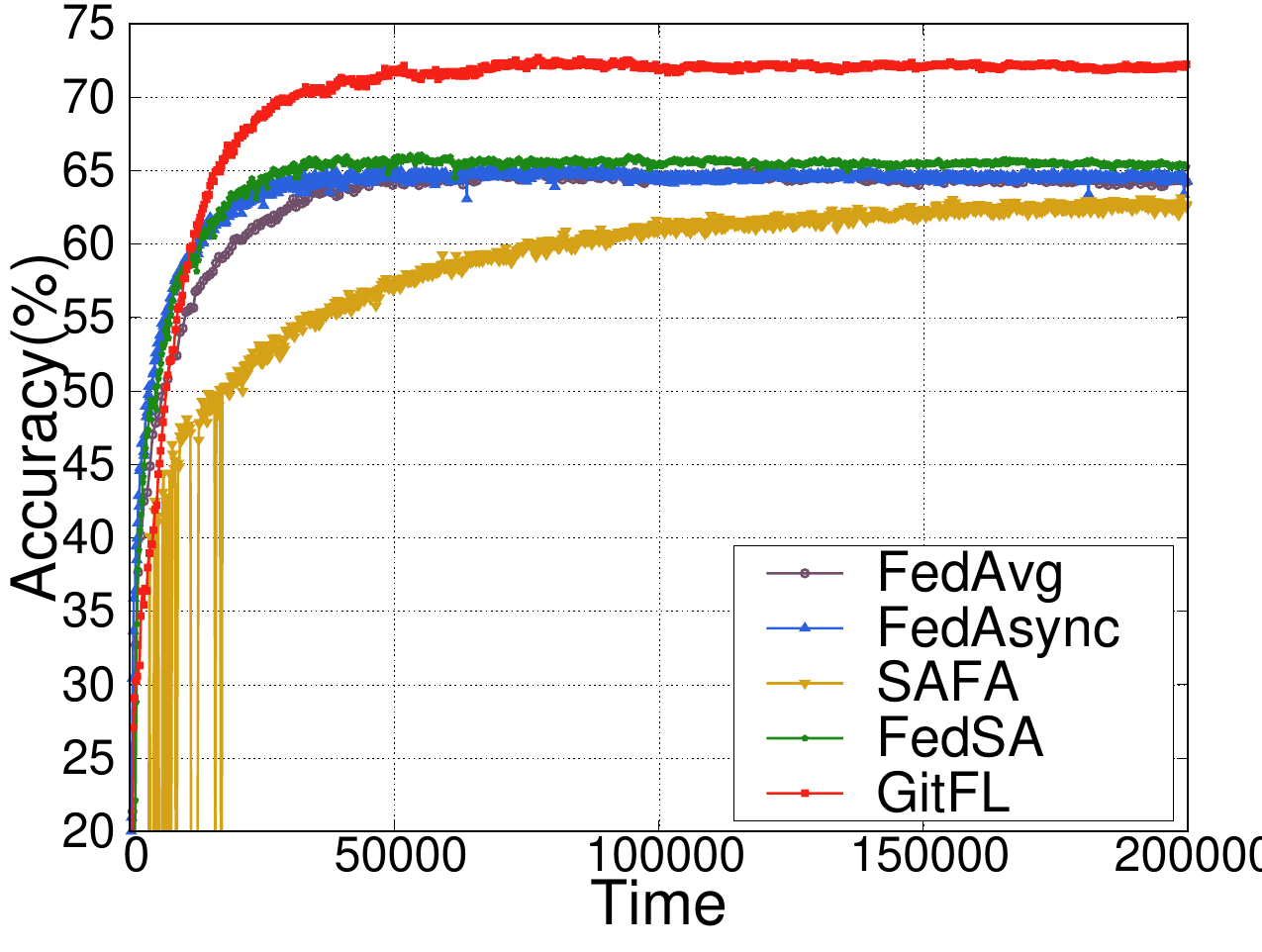}
		\label{fig:femnist}
	}
	\vspace{-0.05in}
	\caption{Learning curves of GitFL and four baseline methods on CIFAR-10}
	\label{fig:accuacy}
	\vspace{-0.05in}
\end{figure}

Figure~\ref{fig:accuacy} presents the trend of  
learning  for  all the FL methods on CIFAR-10 based on ResNet-18 model.
We can observe that a significant  accuracy improvement was made by  GitFL in all  the  non-IID and IID scenarios.
Moreover, we can find that under an  extreme non-IID scenario
with $\alpha=0.1$, the learning curve of 
GitFL is more stable than the ones of all the other baselines.

\subsubsection{Comparison of  Training Time and Communication Overhead}

Given a specific overall FL training target in terms of accuracy, Table~\ref{tab:time} compares the 
training time and communication overhead between GitFL and all the baselines.
From this table, we can find that from the perspective of training time
GitFL outperforms all the baselines in seven out of eight cases. From the perspective of 
communication overhead, GitFL outperforms all the baselines in four
out of eight cases. Note that here all the four cases indicate the higher accuracy targets 
of different settings, respectively.  In other words, when the accuracy target becomes
higher, our approach 
uses much less training time and communication overhead than the other baselines. 
As an example of  dataset CIFAR-10, when $\alpha=0.1$ and the target  
accuracy is 45\%, 
GitFL outperforms FedSA by 2.64X and 1.85X in terms of training time and communication overhead, respectively. Note that in this case, FedSA  has the second-best result.


\subsection{Impacts of Different Configurations}

We checked the  impacts of different configurations on GitFL  from the following 
three perspectives: different compositions of involved devices, 
different number of simultaneously
training  clients, different types of underlying AI models.


\begin{table}[h]
\centering
\vspace{-0.1in}
\caption{Different Configurations of Device Composition}
\vspace{-0.1in}
\scriptsize
\begin{tabular}{|c|c|c|c|c|c|c|}
\hline
\multirow{2}{*}{Configuration} & \multirow{2}{*}{Type} & \multicolumn{5}{c|}{\# of Clients }\\
\cline{3-7}
& & Excellent & High & Medium & Low & Critical\\
\hline
\hline

 \multirow{2}{*}{Config. 1} & Training & 40 & 30 & 10 & 10 & 10\\
 \cline{2-7}
  & Comm. & 40 & 30 & 10 & 10 & 10\\
\hline
\multirow{2}{*}{Config. 2} & Training & 10 & 20 & 40 & 20 & 10\\
\cline{2-7}
  & Comm.& 10 & 20 & 40 & 20 & 10\\
\hline
\multirow{2}{*}{Config. 3} & Training & 10 & 10 & 10 & 30 & 40\\
\cline{2-7}
  & Comm. & 10 & 10 & 10 & 30 & 40\\
\hline
\multirow{2}{*}{Config. 4} & Training & 40 & 10 & 0 & 10 & 40\\
\cline{2-7}
  & Comm. & 40 & 10 & 0 & 10 & 40\\
\hline
\end{tabular}
\label{tab:uncertain}
\end{table}

\subsubsection{Impact of Different Device Compositions}

Table~\ref{tab:uncertain} presents
the four configurations for  different devices used in the experiments, where 
the device settings are provided
in Table~\ref{tab:config}. Note that in each configuration, 
we do not force the devices with better settings 
to be equipped with 
communication modules with higher  quality.

\begin{figure}[h]
\vspace{-0.1in}
	\centering
	\subfigure[Config. 1]{
		\centering
		\includegraphics[width=0.14\textwidth]{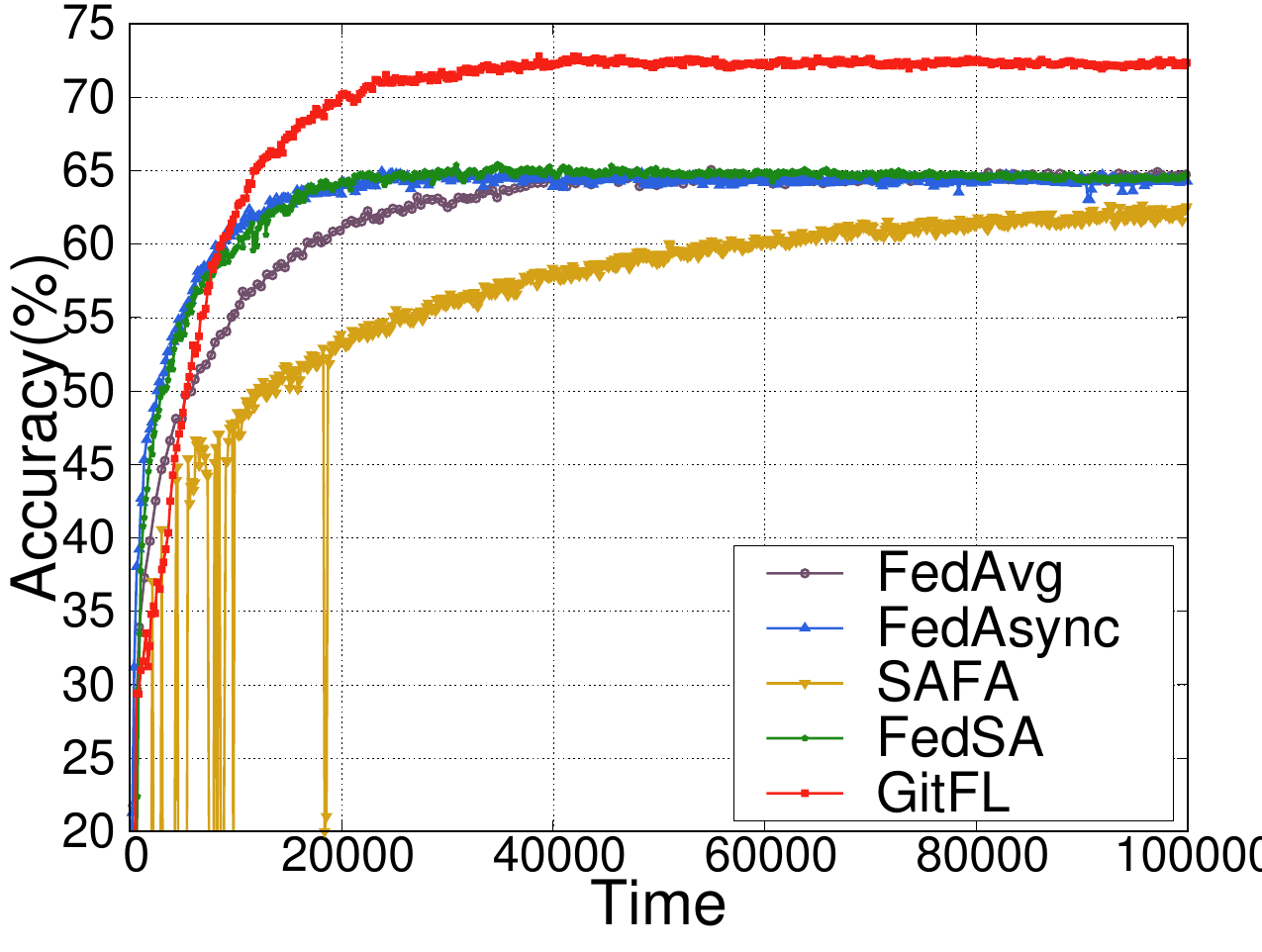}
		\label{fig:high}
	}
	\subfigure[Config. 2]{
		\centering
		\includegraphics[width=0.14\textwidth]{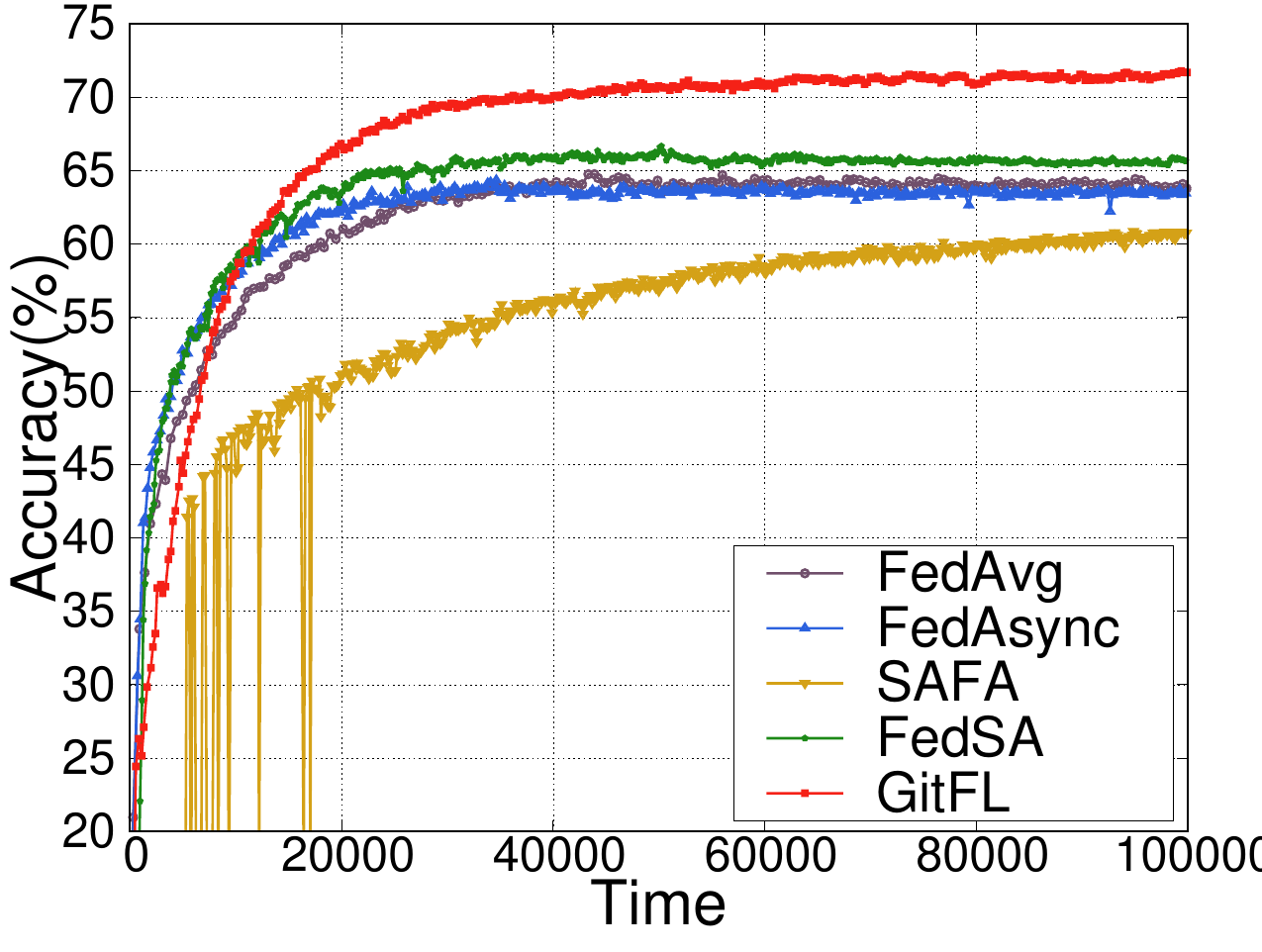}
		\label{fig:medium}
	}
	\\
	\vspace{-0.1in}
	\subfigure[Config. 3]{
		\centering
		\includegraphics[width=0.14\textwidth]{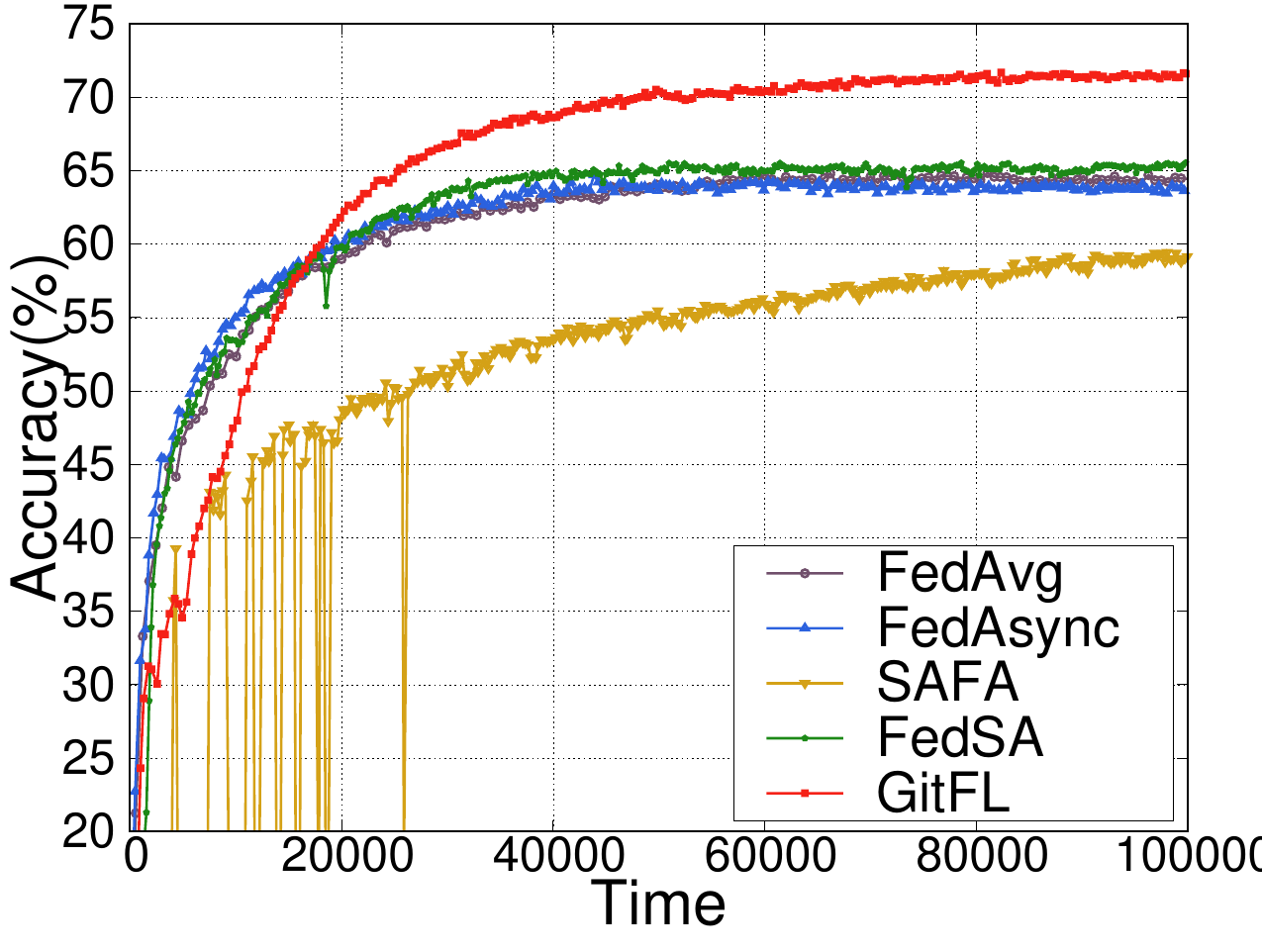}
		\label{fig:low}
	}
        \subfigure[Config. 4]{
		\centering
		\includegraphics[width=0.14\textwidth]{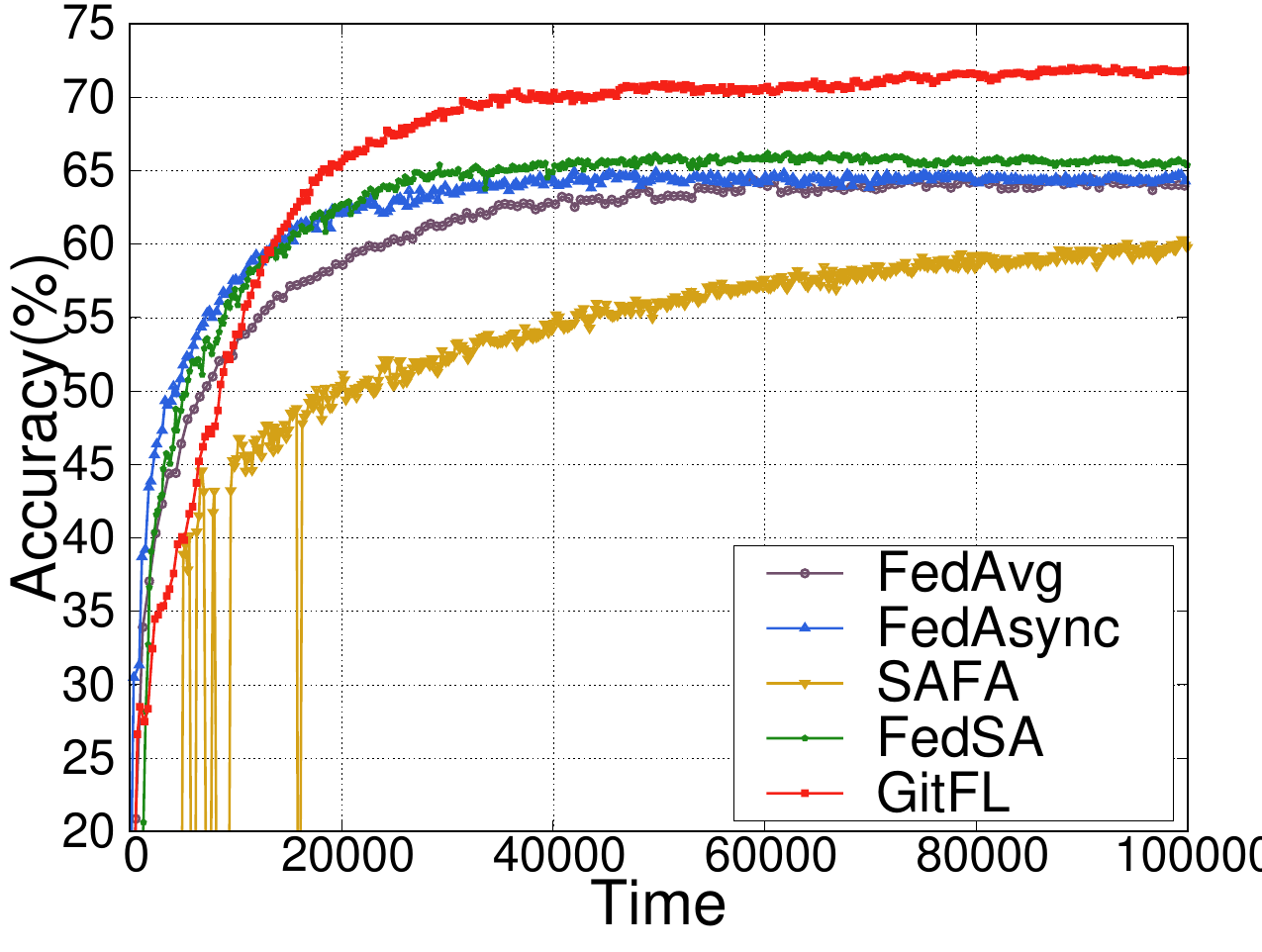}
		\label{fig:low}
	}
	\vspace{-0.05in}
	\caption{Learning curves for different device composition configurations}
	\label{fig:uncertain}
\vspace{-0.05in}
\end{figure}

Figure~\ref{fig:uncertain} presents the 
learning curves of the four device composition configurations, where we   applied  
GitFL (with ResNet-18 model) on dataset CIFAR-10 with IID distribution. 
From Figure~\ref{fig:uncertain}, we can find that 
GitFL achieves 
the highest inference performance for all the four device compositions. 
Note that we can observe similar trends of learning curves
for non-IID scenarios. 

\subsubsection{Impact of Different Number of Simultaneously  Training
Clients}
\begin{figure}[h]
	\vspace{-0.15in}
	\centering
	\subfigure[$K=5$]{
		\centering
		\includegraphics[width=0.14\textwidth]{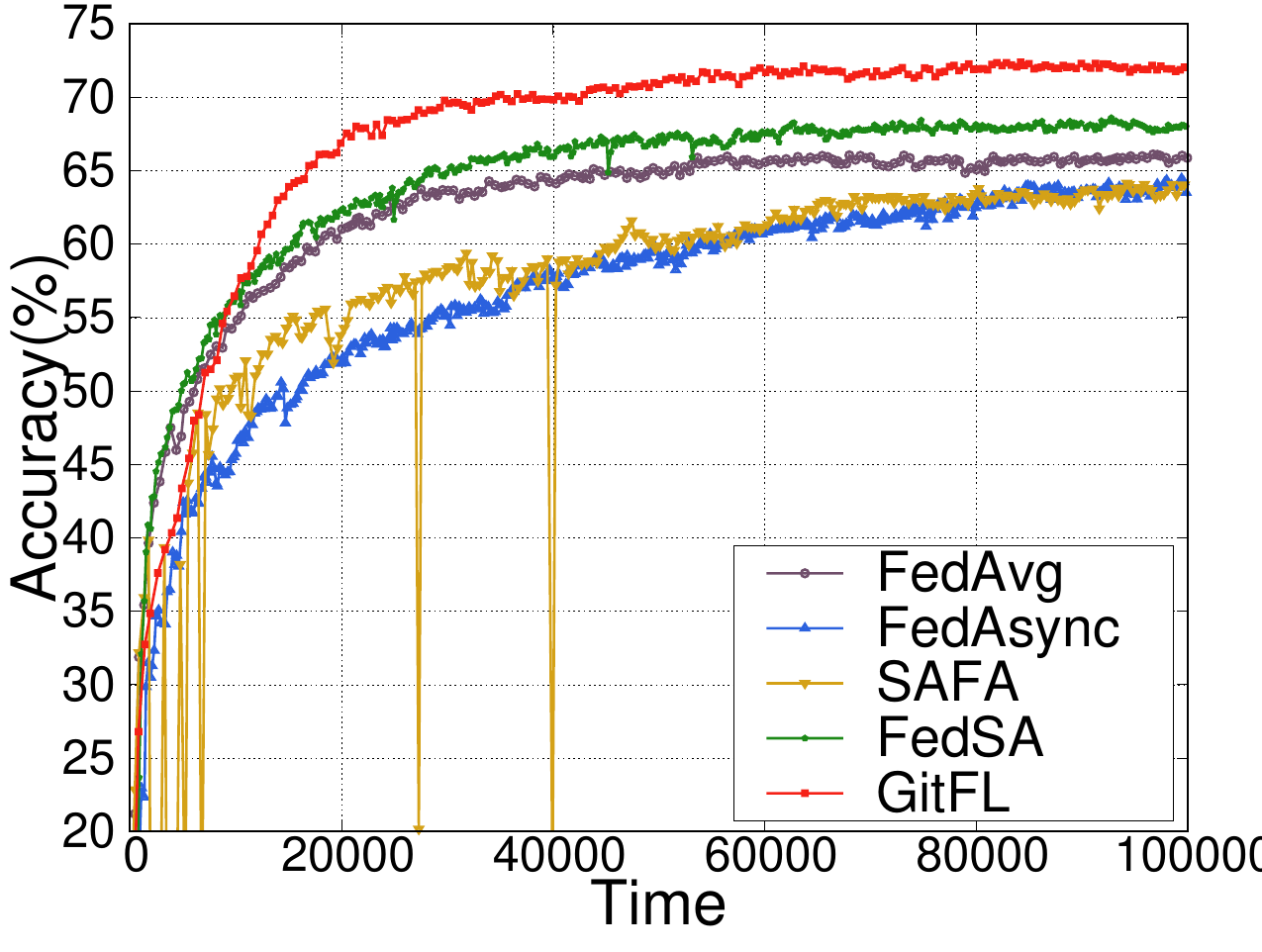}
		\label{fig:100_5}
	}
        \subfigure[$K=10$]{
		\centering
		\includegraphics[width=0.14\textwidth]{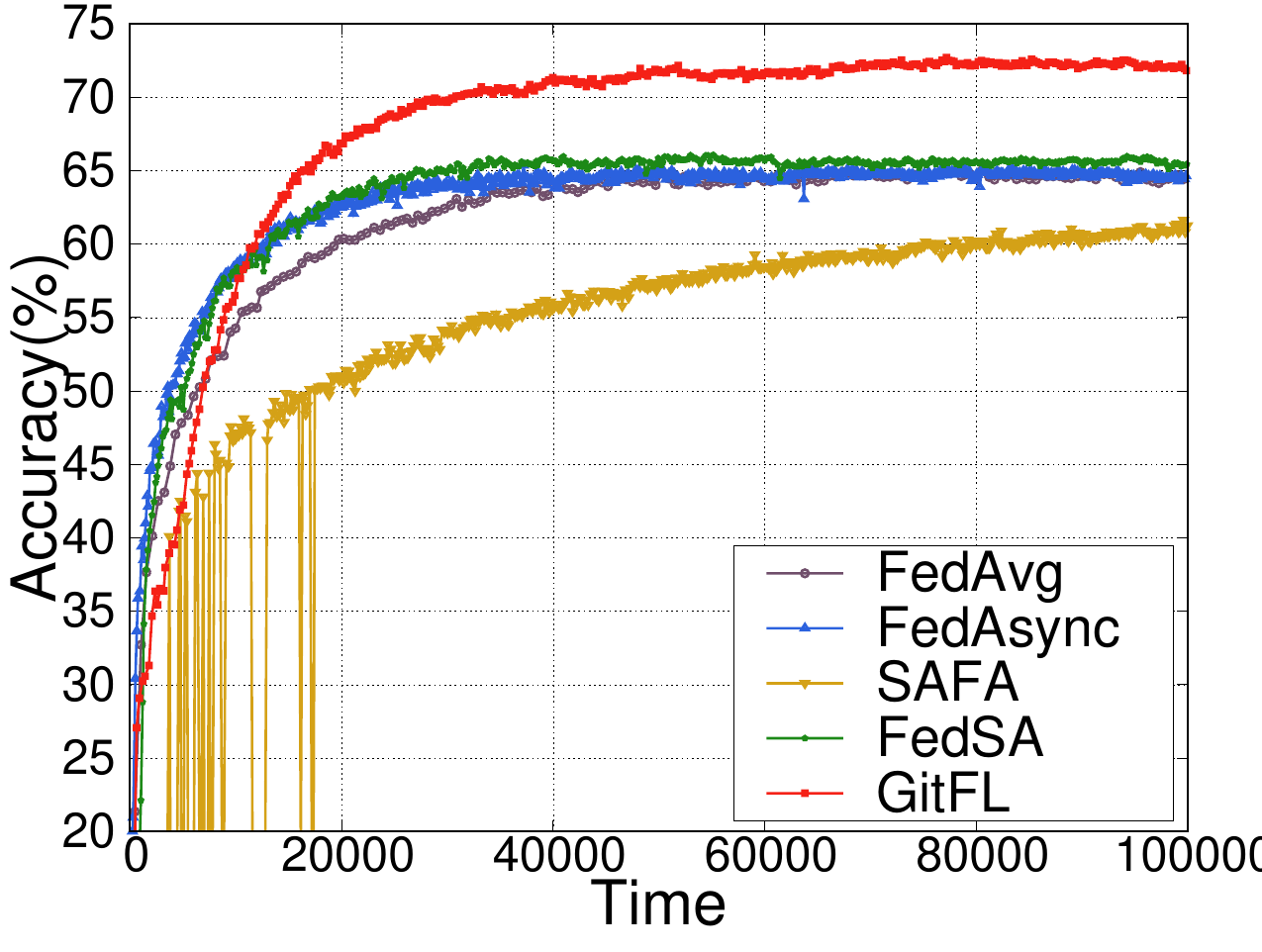}
		\label{fig:100_10}
	}
	\\ \vspace{-0.1 in}
	\subfigure[$K=20$]{
		\centering
		\includegraphics[width=0.14\textwidth]{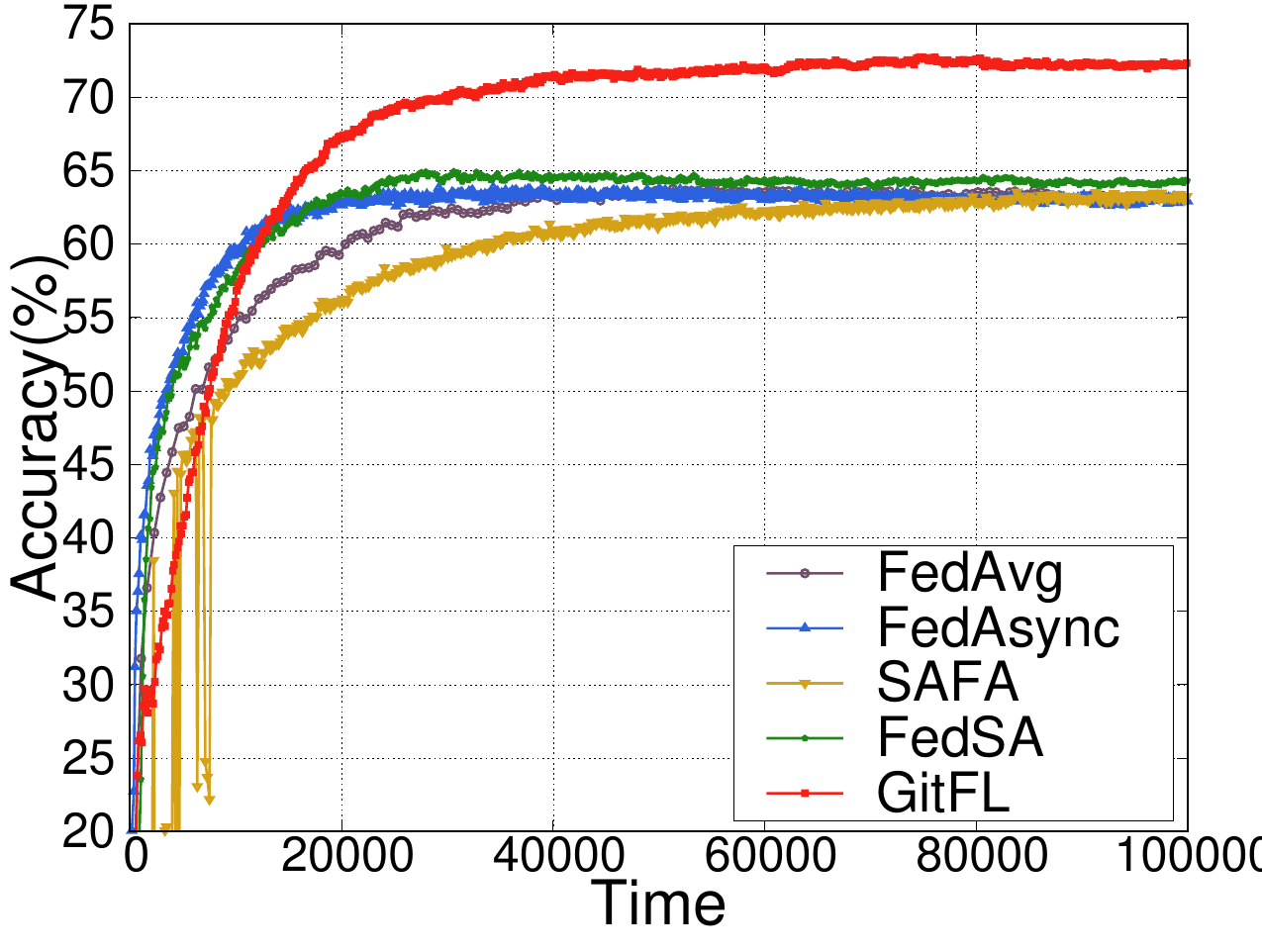}
		\label{fig:100_20}
	}
	\subfigure[$K=50$]{
		\centering
		\includegraphics[width=0.14\textwidth]{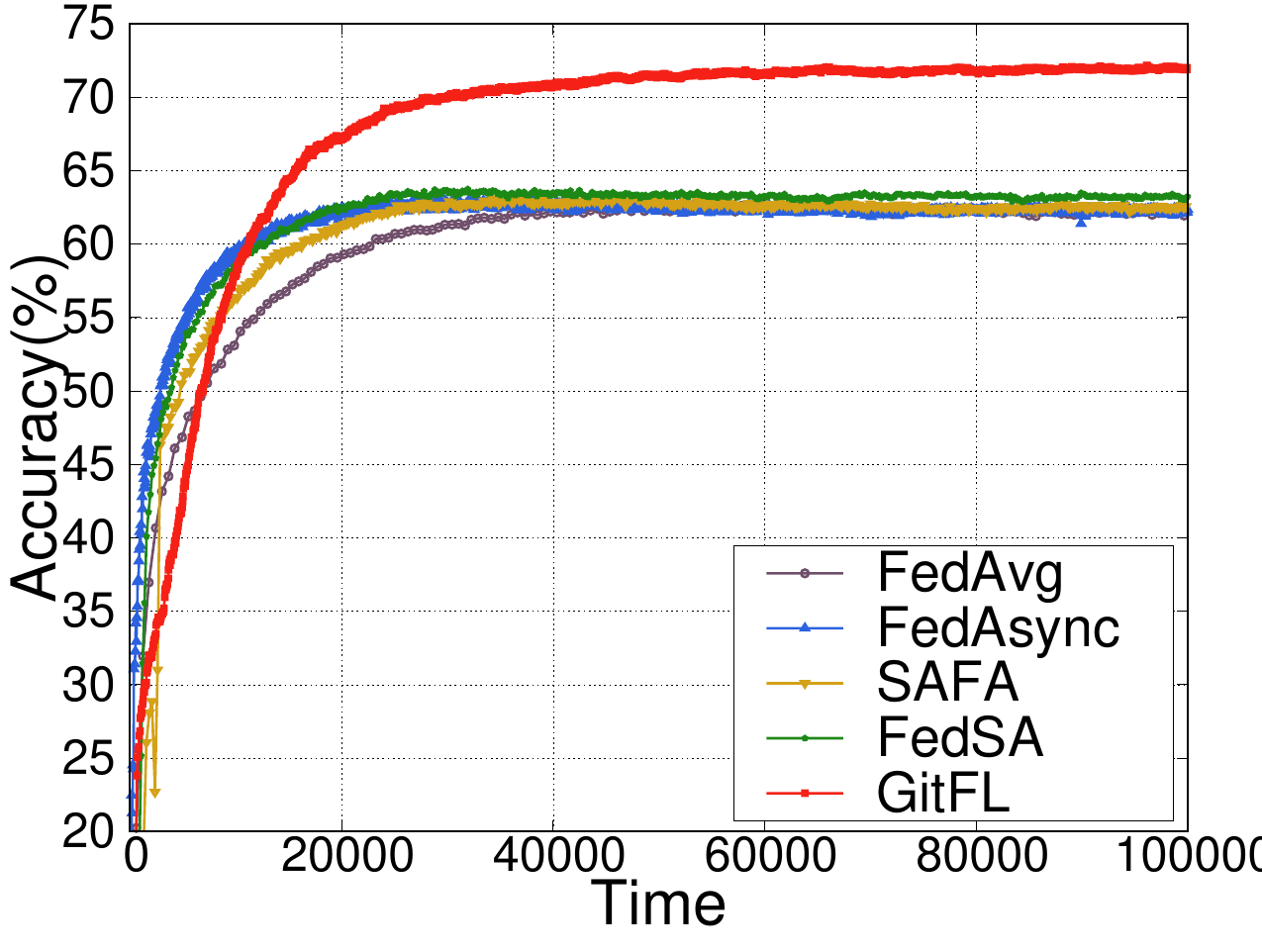}
		\label{fig:100_50}
	}
		\vspace{-0.05in}
	\caption{Learning curves for different number  of simultaneously training clients}
	\vspace{-0.05in}
	\label{fig:client_num}
\end{figure}

To investigate the impacts of the number of simultaneously training clients on GitFL, we considered four cases, where the numbers of simultaneously training clients are 5, 10, 20, and 50, respectively. 
Figure~\ref{fig:client_num} shows  all  experimental results  conducted on CIFAR-10 with IID distribution using ResNet-18.
From this figure, we can find that 
GitFL achieves the highest accuracy with all the settings.
Especially, when more clients are involved in simultaneous training, 
the more improvements GitFL can obtain compared with baseline methods. 

\subsubsection{Impacts of AI Model Types}
\begin{figure}[h]
	\vspace{-0.15 in}
	\centering
	\subfigure[CNN ($\alpha=0.1$)]{
		\centering
		\includegraphics[width=0.14\textwidth]{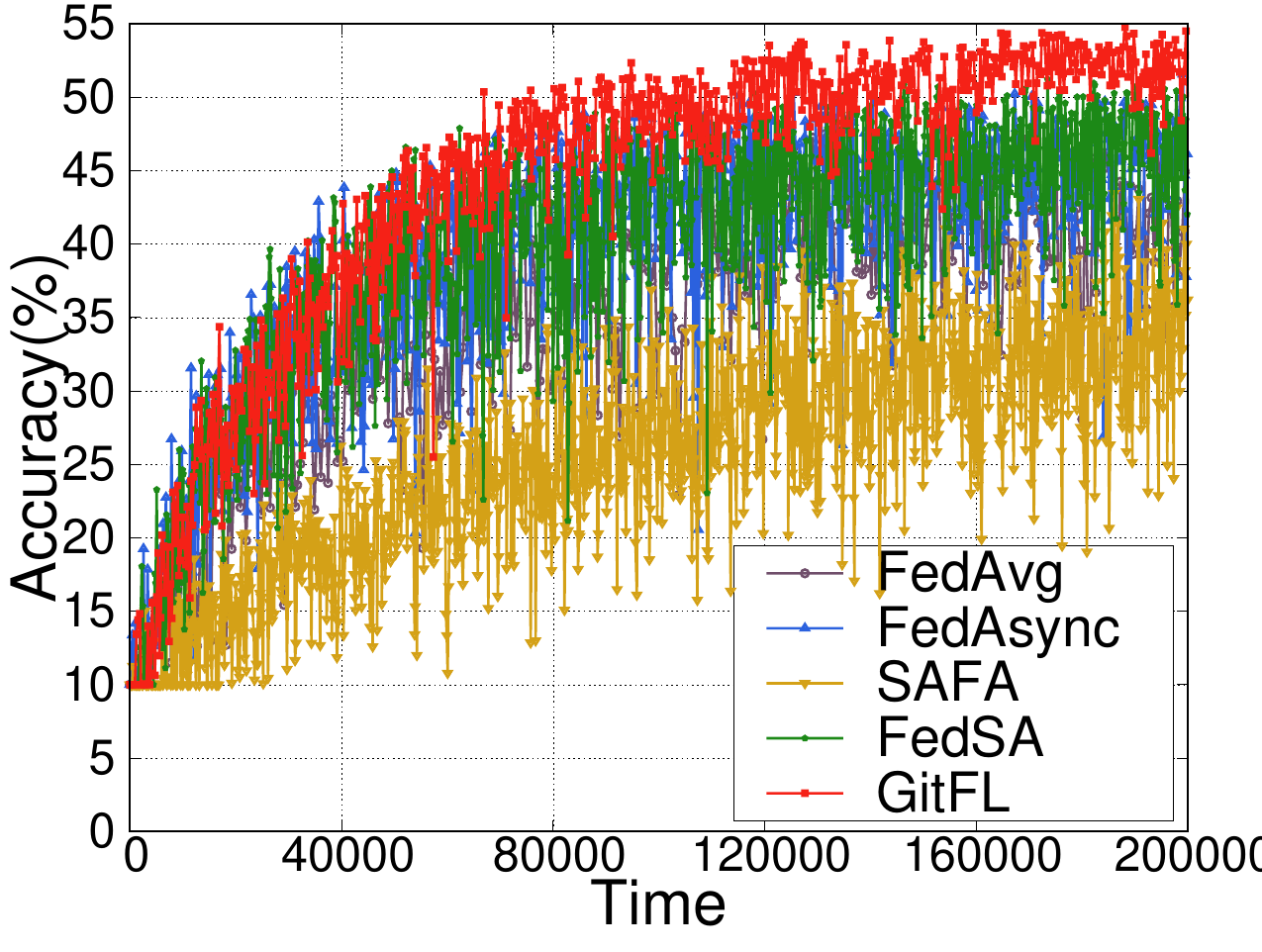}
	}
	\subfigure[CNN (IID)]{
		\centering
		\includegraphics[width=0.14\textwidth]{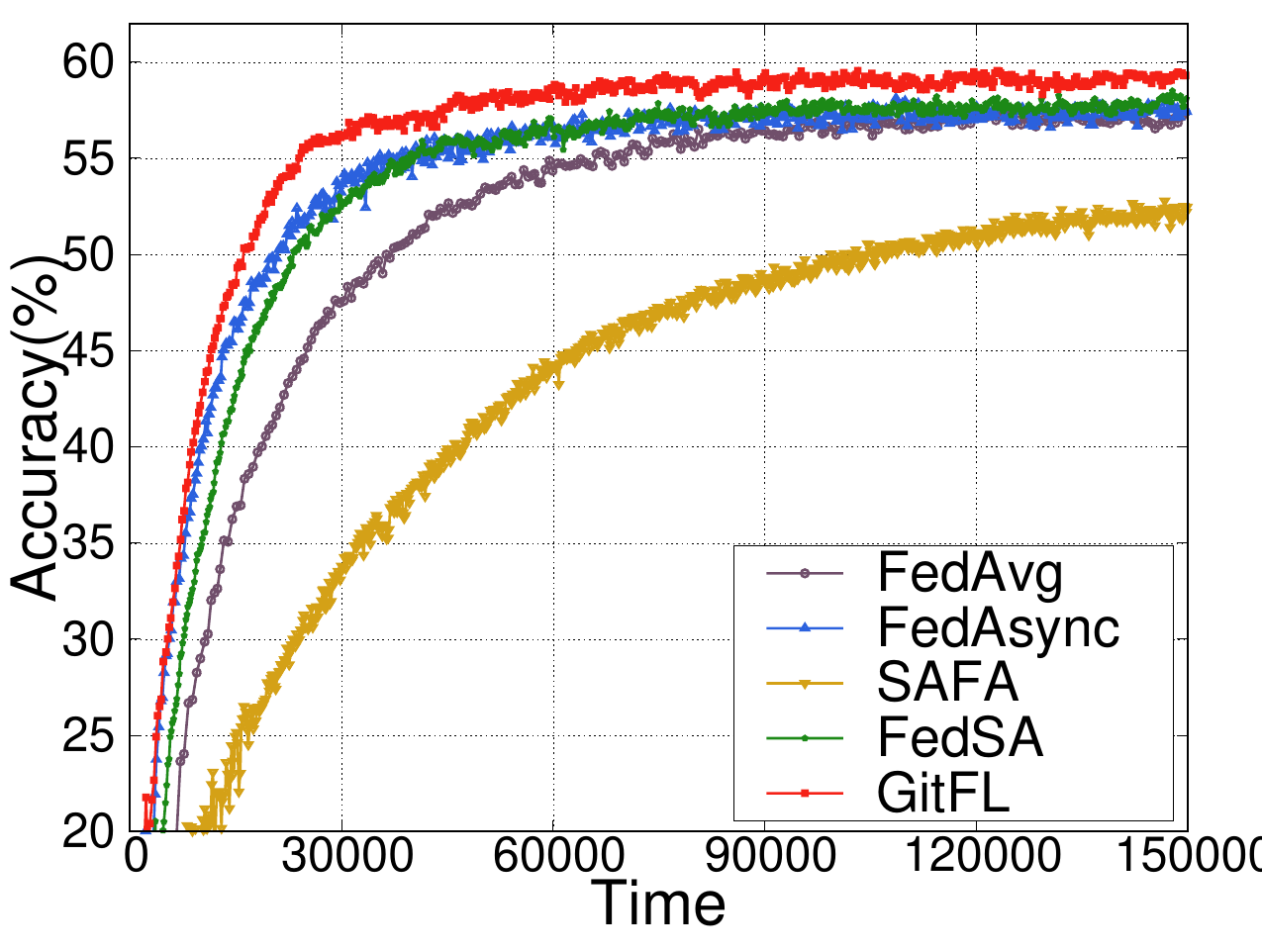}
	}
	\\  	\vspace{-0.1 in}
	\subfigure[VGG-16 ($\alpha=0.1$)]{
		\centering
		\includegraphics[width=0.14\textwidth]{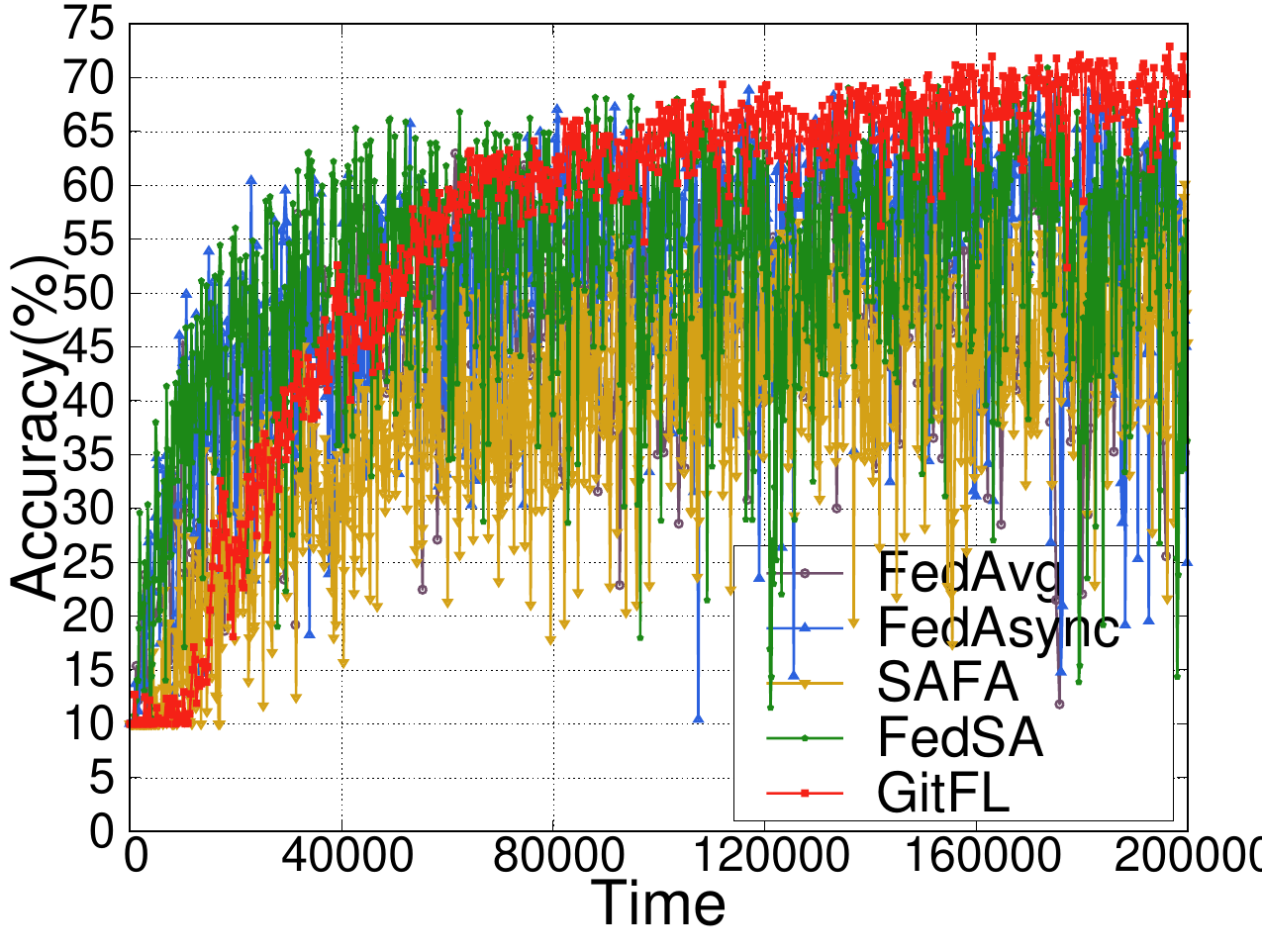}
	}
	\subfigure[VGG-16  (IID)]{
		\centering
		\includegraphics[width=0.14\textwidth]{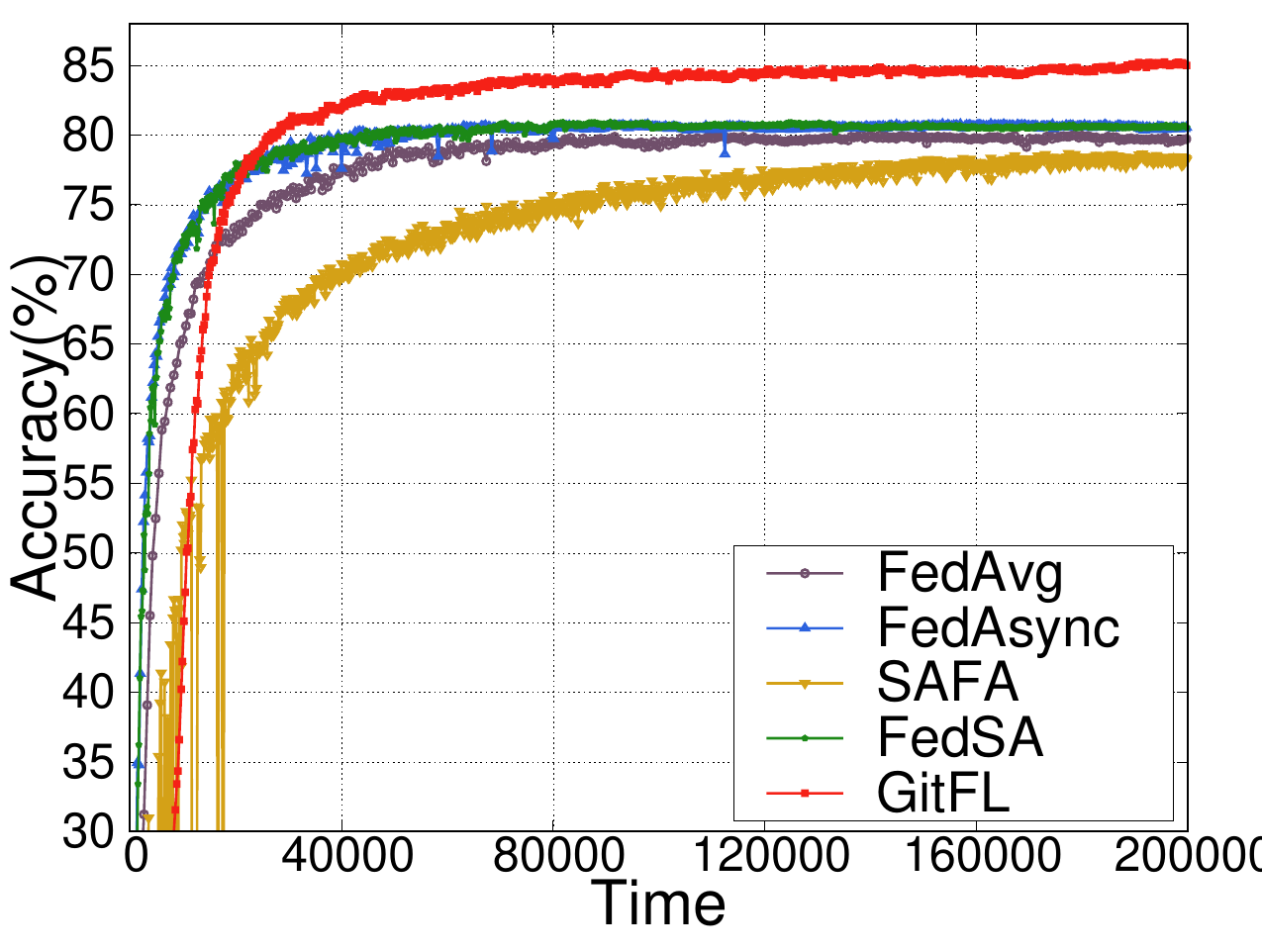}
	}
	\vspace{-0.05 in}
	\caption{Learning curves of GitFL variants on CIFAR-10}
	\label{fig:models}
		\vspace{-0.1 in}
\end{figure}

We investigate two GitFL variants
with different underlying AI models, i.e., CNN and VGG-16. 
Figure~\ref{fig:models} shows the impacts of such two variants with different 
data distributions (i.e., non-IID with $\alpha=0.1$ and IID).
From this figure, we can find that GitFL achieves the highest accuracy in all the cases regardless of underlying AI model types or  data distributions.

\subsection{Ablation Study}
\begin{figure}[h]
		\vspace{-0.1in}
	\begin{center}
		\includegraphics[width=1.3in]{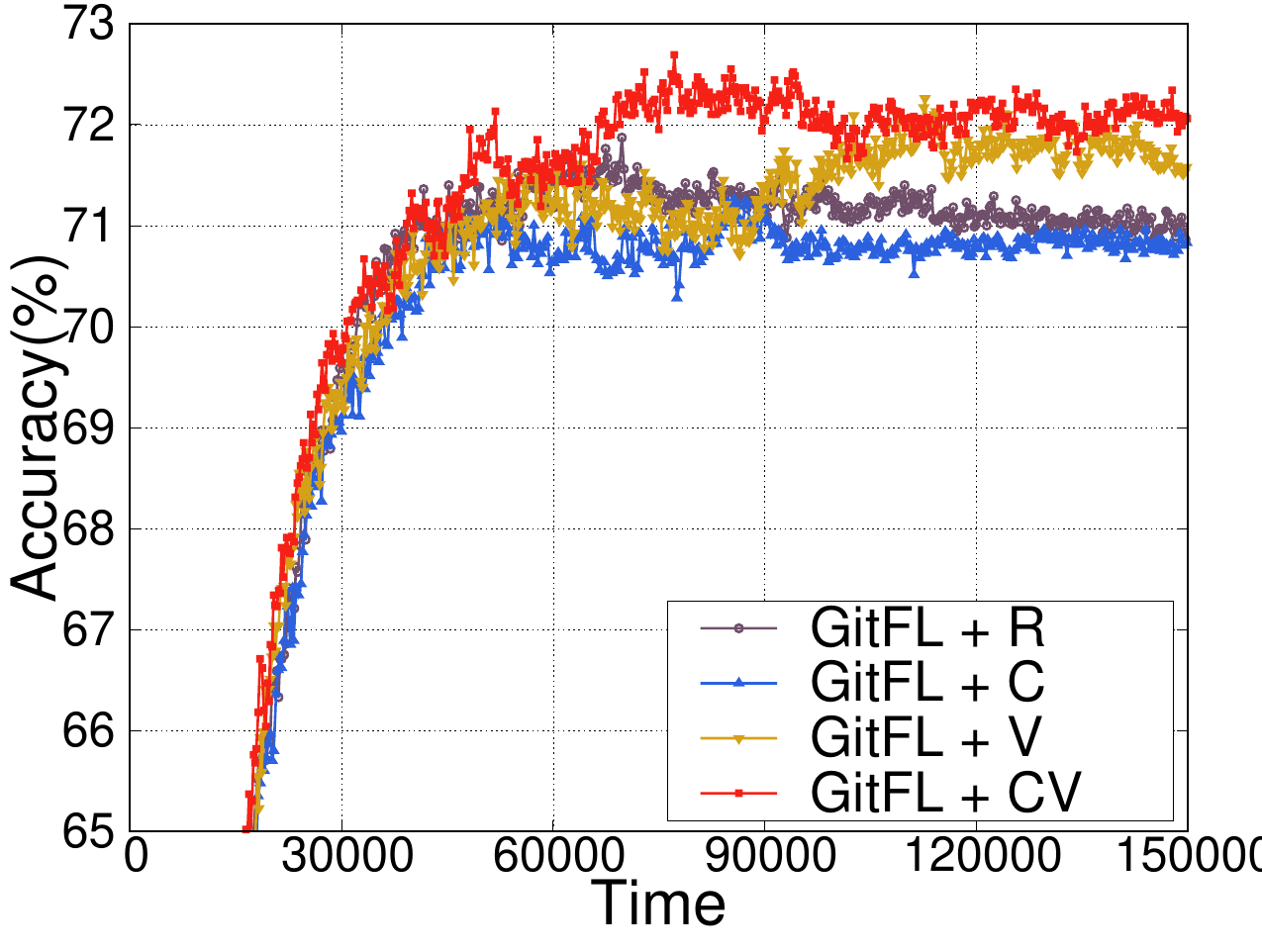}
		\vspace{-0.1in}
		\caption{Ablation study for RL-based client selection} \label{fig:abl}
	\end{center}
	\vspace{-0.2in}
\end{figure}

To demonstrate the effectiveness of our RL-based client selection strategy, we  developed three variants of GitFL: i)
``GitFL+R'' that selects clients for local training in a random manner; 
ii) ``GitFL+C''that selects clients only based on curiosity reward (see Equation \ref{eq:curiosity}); and iii) ``GitFL+V'' that  selects clients only using version reward (see Equation \ref{eq:version}). To facilitate  understanding, we use 
``GitFL+CV'' to indicate the full version of  GitFL implemented in Algorithm 1. 
Figure~\ref{fig:abl} presents the ablation study results on CIFAR-10 dataset
with ResNet-18 following  IID distribution.
We can observe that ``GitFL+CV'' achieves the highest inference performance among 
all the four designs. Note that the improvement of 
 ``GitFL+C''   over  ``GitFL+R'' is negligible, since without 
 the model staleness information the curiosity strategy
 itself cannot benefit   FL
 training.


%% file: conclusion.tex
\section{Conclusion}\label{sec:conclusion}

Due to the notorious
straggler issue, 
it is hard for existing 
FL methods to effectively
control the discrepancies between 
the global model  and delayed local models, resulting in 
low inference accuracy and long training  convergence time.
To address this problem,  
this paper introduced a novel 
asynchronous FL framework named GitFL, whose implementation 
is inspired by the famous version controller system Git.
Unlike traditional FL methods, GitFL aggregates the master model 
based on  both the pushed branch models and their version information. Meanwhile, by adopting  
our proposed RL-based device selection heuristic,  
GitFL supports adaptive 
load balance of versioned 
branch models on devices  according to their staleness. 
Specifically, 
a pulled model  with a  newer
version  will be more likely to 
be assigned on
a less frequently selected
straggler. Comprehensive experimental results show that, 
compared with state-of-the-art asynchronous FL methods, GitFL can 
achieve better FL performance in terms of both 
training time and inference performance considering various 
uncertainties.
